%% file: main.tex
\documentclass[10pt,twocolumn,letterpaper]{article}

\usepackage{cvpr}              

\input{preamble}

\definecolor{cvprblue}{rgb}{0.21,0.49,0.74}
\usepackage[pagebackref,breaklinks,colorlinks,citecolor=cvprblue]{hyperref}
\usepackage{graphicx}
\usepackage{times}
\usepackage{multirow}
\usepackage{epsfig}
\usepackage{amsmath}
\usepackage{amssymb}
\usepackage{booktabs}
\usepackage{verbatim}
\usepackage{subcaption}
\usepackage{adjustbox}
\usepackage{pifont}
\usepackage{enumitem}
\usepackage{comment}
\usepackage{booktabs}
\usepackage{enumitem}
\usepackage{pifont}
\usepackage{algpseudocode}%
\usepackage{listings}%
\usepackage{bbding}
\definecolor{dgreen}{rgb}{0.0,0.6,0.0}
\def\paperID{4374} 
\def\confName{CVPR}
\def\confYear{2024}

\title{LiveHPS: LiDAR-based Scene-level Human Pose and Shape Estimation \\in Free Environment}

\author{
Yiming Ren$^1$, 
Xiao Han$^1$, 
Chengfeng Zhao$^1$,
Jingya Wang$^{1}$,
Lan Xu$^{1}$,
Jingyi Yu$^{1}$,
Yuexin Ma$^{1,}$\footnote[2]{}
\\ 
$^{1}$ ShanghaiTech University\\
{\tt\small \{renym2022,mayuexin\}@shanghaitech.edu.cn}}

\begin{document}

\makeatletter
\let\@oldmaketitle\@maketitle
\renewcommand{\@maketitle}{
   \@oldmaketitle
 \begin{center}
    \vspace{-6ex}
      \includegraphics[width=1.0\linewidth]{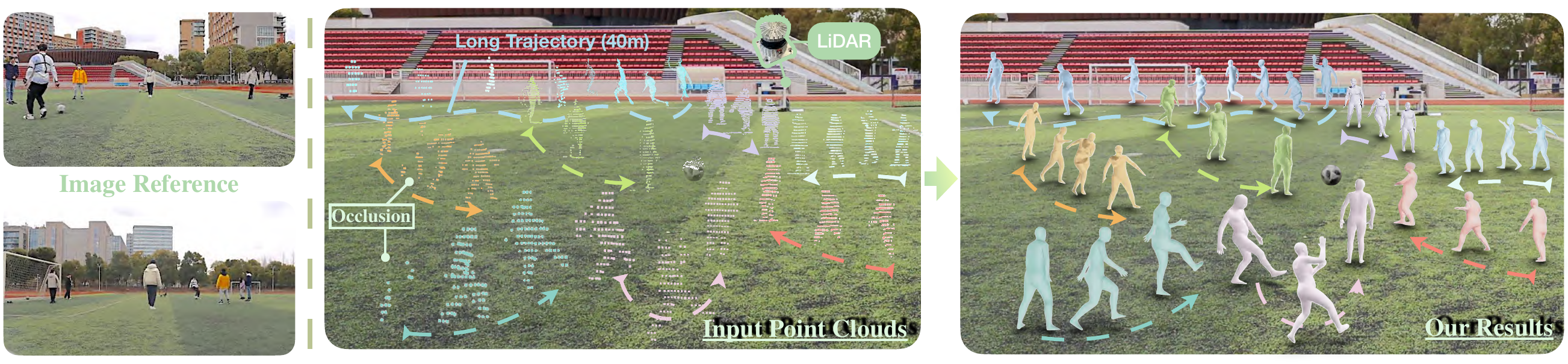}
 \end{center}
 \vspace{-2ex}
  \refstepcounter{figure}\normalfont Figure~\thefigure. 
  We propose a novel single-LiDAR-based approach for 3D HPS in large-scale scenarios, which is not limited for fixed studios, light conditions, and wearable devices. Our method predicts full human SMPL parameters(pose, shape, translation) from consecutive LiDAR point clouds and performs well for challenging poses and occlusion situations.
  \label{fig:teaser}
  \newline
  }

\maketitle
\input{sections/Abstract}    
\input{sections/Intro}

\input{sections/Related_work}
\input{sections/Method}
\input{sections/Dataset}

\input{sections/Experiments}
\input{sections/Conclusion}

\section{Acknowledgements}
This work was supported by NSFC (No.62206173), Shanghai Sailing Program (No.22YF1428700), Natural Science Foundation of Shanghai (No.22dz1201900), MoE Key Laboratory of Intelligent Perception and Human-Machine Collaboration (ShanghaiTech University), Shanghai Frontiers Science Center of Human-centered Artificial Intelligence (ShangHAI).

{
    \small
    \bibliographystyle{ieeenat_fullname}
    \bibliography{main}
}

\input{sections/sup}

\end{document}

%% file: preamble.tex
\usepackage[dvipsnames]{xcolor}


%% file: sections/Abstract.tex
\begin{abstract}
For human-centric large-scale scenes, fine-grained modeling for 3D human global pose and shape is significant for scene understanding and can benefit many real-world applications. In this paper, we present \textbf{LiveHPS}, a novel single-\underline{\textbf{Li}}DAR-based approach for scene-le\underline{\textbf{ve}}l \underline{\textbf{H}}uman \underline{\textbf{P}}ose and \underline{\textbf{S}}hape estimation without any limitation of light conditions and wearable devices. In particular, we design a distillation mechanism to mitigate the distribution-varying effect of LiDAR point clouds and exploit the temporal-spatial geometric and dynamic information existing in consecutive frames to solve the occlusion and noise disturbance. LiveHPS, with its efficient configuration and high-quality output, is well-suited for real-world applications.
Moreover, we propose a huge human motion dataset, named \textbf{FreeMotion}, which is collected in various scenarios with diverse human poses, shapes and translations. It consists of multi-modal and multi-view acquisition data from calibrated and synchronized LiDARs, cameras, and IMUs. Extensive experiments on our new dataset and other public datasets demonstrate the SOTA performance and robustness of our approach. We will release our code and dataset soon.
\end{abstract}

\vspace{-6ex}

%% file: sections/Intro.tex
\section{Introduction}

Human pose and shape estimation (HPS) is aimed at reconstructing 3D digital representations of human bodies, such as SMPL~\cite{SMPL2015}, using data captured by sensors. It is significant for two primary applications: one in motion capture for the entertainment industry, including film, augmented reality, virtual reality, mixed reality, etc.; and the other in behavior understanding for the robotics industry, covering domains like social robotics, assistive robotics, autonomous driving, human-robot interaction, and beyond.

While optical-based methods~\cite{DeepCap_CVPR2020,challencap,PHMR_ICCV2021,HUMOR_ICCV2021,PARE_ICCV2021} have seen significant advancements in recent years, their efficacy is limited due to the camera sensor's inherent sensitivity to variations in lighting conditions, rendering them impractical for use in uncontrolled environments. In contrast, inertial methods~\cite{XSENS,noitom,yi2021transpose,PIPCVPR2022} utilize body-mounted inertial measurement units (IMUs) to derive 3D poses, which is independent of lighting and occlusions. However, these methods necessitate the use of wearable devices, struggle with drift issues over time, and fail to capture human body shapes and precise global translations.

LiDAR is a commonly used perception sensor for robots and autonomous vehicles~\cite{zhu2021cylindrical,zhu2020ssn,cong2022stcrowd} due to its accurate depth sensing without light interference. Recent advances~\cite{ren2023lidar} in HPS are turning to utilize LiDAR for capturing high-quality SMPLs in the wild. LiDARCap~\cite{li2022lidarcap} proposes a GRU-based approach for estimating only human pose parameters from LiDAR point clouds. MOVIN~\cite{jang2023movin} uses a CVAE framework to link point clouds with human poses for both human pose and global translation estimation. Nevertheless, these approaches lack the capability to estimate body shapes, and further, they disregard the challenging characteristics of LiDAR point clouds, leading to an unstable performance in real-world scenarios. First, the distribution and pattern of LiDAR point clouds vary across different capture distances and devices. Second, the view-dependent nature of LiDAR results in incomplete point clouds of the human body, affected by self-occlusion or external obstruction. Third, real-captured LiDAR point clouds invariably contain noise in complex scenarios, caused by the reflection interference or carry-on objects. These properties all bring challenges for accurate and robust HPS in extensive, uncontrolled environments.


Considering above intractable problems of LiDAR point cloud, we introduce LiveHPS, a novel single-LiDAR-based approach for capturing high-quality human pose, shape, and global translation in large-scale free environment, as shown in Fig.~\ref{fig:teaser}. The deployment-friendly single-LiDAR setting is unrestricted in acquisition sites, light conditions, and wearable devices, which can benefit many practical applications.
In order to improve the robustness for tackling point distribution variations, we design an \textbf{Adaptive Vertex-guided Distillation} module to make diverse point distributions align with the regular SMPL mesh vertex distribution in high-level feature space by a prior consistency loss. Moreover, to reduce the influence of occlusion and noise, we propose a \textbf{Consecutive Pose Optimizer} to explore the geometric and dynamic information existing in temporal and spatial spaces for pose refinement by attention-based feature enhancement. In addition, a \textbf{Skeleton-aware Translation Solver} is also presented to eliminate the effect of incomplete and noised point cloud on accurate estimation for human global translation. In particular, we introduce the scene-level unidirectional Chamfer distance (SUCD) from the input point cloud to the estimated human mesh vertex in global coordinate system as a new evaluation measurement for LiDAR-based HPS, which can reflect the fine-grained geometry error and translation error between the prediction and the ground truth.

It is worth noting that we also introduce \textbf{FreeMotion}, a novel huge motion dataset captured in diverse large-scale real scenarios with multiple persons, which contains multi-modal data(LiDAR point clouds, RGB image and IMUs), multi-view data(front, back and side), and comprehensive SMPL parameters(pose, shape and global translation). Through extensive experiments and ablation studies on FreeMotion and other public datasets, our method outperforms others by a large margin. 

Our main contributions can be summarized as follows:
\begin{itemize}
    \item We present a novel single-LiDAR-based method for 3D HPS in large-scale free environment, which achieves state-of-the-art performance. 
    \item We propose an effective vertex-guided adaptive distillation module, consecutive pose optimizer, and skeleton-aware translation solver to deal with the distribution-varied, incomplete, and noised LiDAR point clouds.
    \item We present a new motion dataset captured in diverse real scenarios with rich modalities and annotations, which can facilitate further research of in-the-wild HPS.
\end{itemize}

%% file: sections/Related_work.tex
\section{Related Work}
\subsection{Optical-based Methods}
Optical motion capture technology has advanced from initial marker-based systems~\cite{VICON,Vlasic2007,optitrack} that rely on camera-tracked markers to reconstruct a 3D mesh, to markerless systems~\cite{JooLTGNMKNS2015,luo2021dynamics,AminARS2009,BurenSC2013,ElhayAJTPABST2015,RhodiRRST2015,Robertini:2016,Pavlakos17,Simon17} which track human motion using cameras through feature detection and matching algorithms. Despite they can get high-accuracy results, these systems are often expensive and require elaborate setup and calibration. 
To mitigate these challenges, monocular mocap methods using optimization~\cite{TAM_3DV2017,Lassner17,bogo2016keep,Kolotouros_2019_CVPR} and regression~\cite{HMR18,Kanazawa_2019CVPR,VIBE_CVPR2020,zanfir2020neural}, along with template-based, probabilistic~\cite{MonoPerfCap,LiveCap2019tog,EventCap_CVPR2020,DeepCap_CVPR2020,challencap}, and semantic-modeling techniques~\cite{PARE_ICCV2021}, have emerged to address monocular system limitations. Nonetheless, these approaches still suffer from inherent light sensitivity and depth ambiguity.
Some strategies\cite{Shotton:2011,Baak:2011,Wei:2012,DoubleFusion,guoTwinFusion} incorporate depth cameras to resolve depth ambiguity, yet these cameras have a limited sensing range and are ineffective in outdoor scenes.

\subsection{Inertial-based Methods}
Unlike optical systems, inertial motion capture systems~\cite{XSENS} are not affected by light conditions and occlusions. They generally need numerous IMUs attached to form-fitting suits, a setup that can be heavy and inconvenient, motivating interest in more sparse configurations, such as six-IMU setting~\cite{von2017SIP,huang2018DIP,yi2021transpose,PIPCVPR2022} and four-IMU setting~\cite{ren2023lidar}. 
However, these methods suffer from drift errors over time, cannot provide precise shape and global translation, and require wearable devices, not practical for daily-life scenarios.


\subsection{LiDAR-based Methods}
With precise long-range depth-sensing ability, LiDAR has emerged as a key sensor in robotics and autonomous vehicles~\cite{Cong_2022_CVPR,zhu2020ssn,zhu2021cylindrical,Yin2020Centerbased3O,peng2022cl3d}. LiDAR can provide precise depth information and global translation in expansive environments, remaining uninfluenced by lighting conditions, enabling robust 3D HPS. Recently, PointHPS~\cite{cai2023pointhps} provides a cascaded network architecture for pose and shape estimation from point clouds. However, it is applicable for dense point clouds rather than sparse LiDAR point clouds. LiDARCap~\cite{li2022lidarcap} employs a graph-based convolutional network to predict daily human poses in LiDAR-captured large-scale scenes. MOVIN~\cite{jang2023movin} presents a generative method for estimating both pose and global translation. However, these methods cannot predict full SMPL parameters (pose, shape, and global translation) and are fragile for complex real scenarios with occlusion and noise.

\subsection{3D Human Motion Datasets}
Data-driven 3D HPS have gained traction in recent years benefiting from extensive labeled datasets. Indoor marker-based datasets like Human3.6M~\cite{ionescu2013human3} and HumanEva~\cite{SigalBB2010} use multi-view camera systems to record daily motions. AMASS~\cite{AMASS_ICCV2019} unifies these datasets, providing a standardized benchmark for network training. Marker-less datasets such as MPI-INF-3DHP~\cite{mehta2017monocular} and AIST++~\cite{li2021learn} capture more complex poses without constraint of the wearable devices, all above datasets are still confined to indoor settings.
Outdoor motion capture datasets like PedX~\cite{kim2019pedx} and 3DPW~\cite{von2018recovering} capture motions in the wild but lack accurate depth information, hindering scene-level human motion research. HuMMan~\cite{cai2022humman} constitutes a mega-scale database that offers high-resolution scans for subjects, and MOVIN~\cite{jang2023movin} provides motion data from multi-camera capture system with point clouds, but both datasets are limited in short-range scenes. ~\cite{dai2022hsc4d, dai2023sloper4d, yan2023cimi4d} are proposed for human motion capture in large-scale scenes using environment-involved optimization, but they are limited in a single-person setting. Recently, LiDARHuman26M~\cite{li2022lidarcap} and LIP~\cite{ren2023lidar} provide LiDAR-captured motion dataset in large scenes, but both datasets exclusively provide pose parameters of SMPL in single-person scenarios. 
In contrast, we propose a large-scale LiDAR-based motion dataset with full SMPL parameter annotations. It comprises a variety of challenging scenarios with occlusions and interactions among multiple persons and objects, which has great practical significance.

%% file: sections/Method.tex
\newcommand{\rot}[2]{\mathbf{R}_{#1}^{#2}}
\newcommand{\acc}[2]{\mathbf{a}_{#1}^{#2}}

\section{Methodology}

\begin{figure*}[ht]
	\centering
 \vspace{-1ex}
	\includegraphics[width=\linewidth]{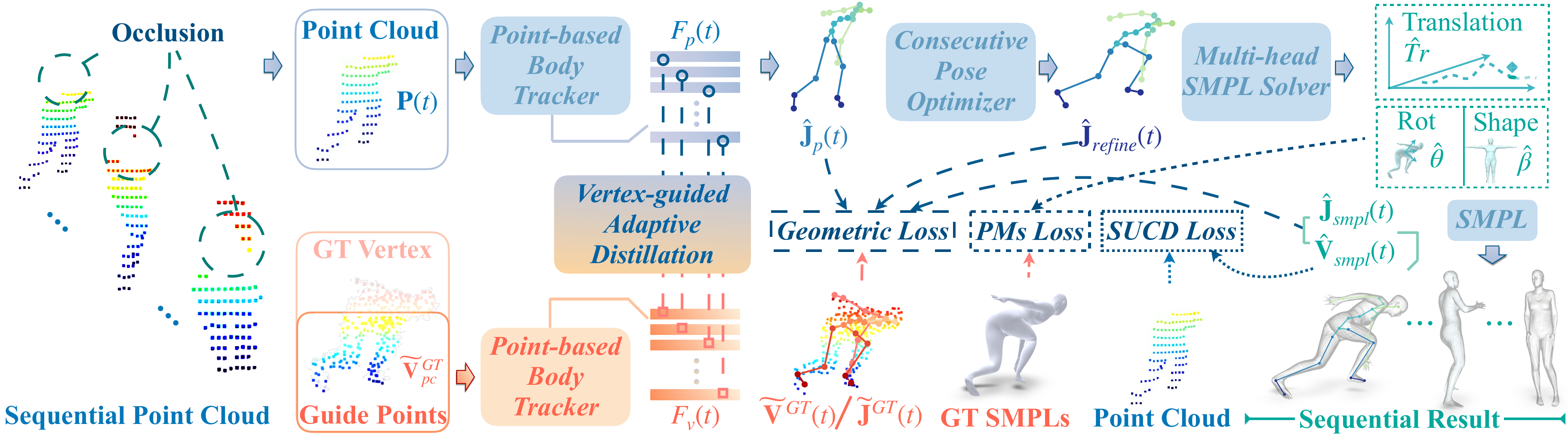}
 \vspace{-4ex}
	\caption{The pipeline of LiveHPS. With sequential LiDAR point clouds as input, LiveHPS consists of three critical modules to obtain human SMPL parameters, including a point-based body tracker to distill the pose-prior information, a consecutive pose optimizer to refine the pose via utilizing joint-wise features, and a multi-head SMPL solver to regress parameters of human models. }
	\label{fig:pipeline}
	\vspace{-2ex}
\end{figure*}

We propose a single-LiDAR-based approach named LiveHPS for scene-level 3D human pose and shape estimation in large-scale free environments. 
The overview of our pipeline is shown in Fig.~\ref{fig:pipeline}. We take consecutive 3D single-person point clouds as input and aim to acquire consistently accurate local pose, human shape, and global translation without any limitation of acquisition sites, light conditions, and wearable devices. There are three main procedures in our network, including point-based body tracker (Sec.~\ref{sec:body_tracker}), consecutive pose optimizer (Sec.~\ref{sec:pose_optimizer}), and attention-based multi-head SMPL solver (Sec.~\ref{sec:IK}). First, we utilize the point-based body tracker to extract point-wise features and predict the human body joint positions. Second, we propose the attention-based temporal-spatial feature enhancement mechanism to acquire refined joint positions using joint-wise geometric and relationship features. Finally, we design an attention-based multi-head solver to regress the human SMPL parameters including human local pose, shape and global translation from the refined body skeleton. 

\subsection{Preliminaries}

LiveHPS takes a consecutive sequence of single-person point clouds with $T$ frames as input. As raw point clouds have various numbers of points at different times $t$, we implement normalization process by resampling each frame to a fixed $N_{fps}=256$ points utilizing the farthest point sampling algorithm (FPS) and subtracting the average locations $\mathbf{loc}(t)\in \mathbb{R}^{3} $ of the raw data. $\mathbf{P}(t)\in \mathbb{R}^{3N_{fps}}$ denotes the pre-processed input at time $t$.

We define $N_J$ as the number of body joints and $N_V$ as the number of body vertices on SMPL mesh;  $\hat{\mathbf{J}}(t),\ \widetilde{\mathbf{J}}^{GT}(t)\in \mathbb{R}^{3N_J}$ as predicted and ground-truth root-relative joint positions at time $t$, repectively; $\hat{\mathbf{V}}(t), \widetilde{\mathbf{V}}^{GT}(t)\in \mathbb{R}^{3N_V}$ as predicted and ground-truth vertex positions. Our network prediction consists of
$\hat{\mathbf{\theta}}(t)\in \mathbb{R}^{6N_J}, \hat{\mathbf{\beta}}\in \mathbb{R}^{10}$ and $\hat{Tr}(t)\in \mathbb{R}^{3}$, the pose, shape, and global translation parameters of SMPL. $\mathbf{\theta}^{GT}(t), \mathbf{\beta}^{GT}$ and $Tr^{GT}(t)$ are corresponding ground truth. We use 6D-rotation-based pose representation.

\subsection{Point-based Body Tracker}
\label{sec:body_tracker}

For the input of our pre-processed consecutive point clouds, we extract the point-wise feature following the PointNet-GRU structure proposed by LIP~\cite{ren2023lidar} and regress the human body joint positions with an MLP decoder. Considering the irregular distribution of LiDAR point clouds vary across different capture distances and devices, and are also effected by occlusion and noise (Fig.~\ref{fig:teaser}), we design a \textbf{Vertex-guided Adaptive Distillation (VAD)} mechanism to unify the point distribution to facilitate the training of the network and improve the robustness. Because the vertices of SMPL mesh have relatively regular representation, we make diverse point distributions aligned with the mesh vertex distribution in high-level feature space by distillation, as Fig.~\ref{fig:pipeline} shows. 

Firstly, we use the global translation $Tr^{GT}(t)$ to align the LiDAR point cloud $\mathbf{P}(t)$ with the ground truth mesh vertex $\widetilde{\mathbf{V}}^{GT}(t)$ and utilize k-Nearest-Neighbours (kNN) algorithm to sample the corresponding vertices, defined as $\widetilde{\mathbf{V}}^{GT}_{pc}(t)$. Then, we use $\widetilde{\mathbf{V}}^{GT}_{pc}(t)$ to pre-train a vertex body tracker to regress the joint positions $\hat{\mathbf{J}}_{v}(t)$. We use the mean squared error (MSE) loss ${L}_{mse}(\hat{\mathbf{J}}_{v})$ for supervision:
\begin{equation}
    \begin{aligned}
    \mathcal{L}_{mse}(\hat{\mathbf{J}}_{v})= \sum_t \parallel \hat{\mathbf{J}}_{v}(t)-\widetilde{\mathbf{J}}^{GT}(t) \parallel_2^2.
    \end{aligned}
\label{equ:loss_Vertex}
\end{equation}
Subsequently, we input sequential point clouds $\mathbf{P}(t)$ and their corresponding vertex data $\widetilde{\mathbf{V}}^{GT}_{pc}(t)$ into two independent body trackers to obtain point-wise features $F_p(t)\in \mathbb{R}^{k}$ and $F_v(t)\in \mathbb{R}^{k}$, respectively, where $k = 1024$. Notably, two point-based body tracker networks share distinct weights and we freeze the pre-trained parameters of the vertex body tracker during training. 
To align real point distributions with the regular vertex distribution, we employ a pose-prior consistency loss $\mathcal{L}_{pc}$ to minimize the high-level feature distance between LiDAR point clouds and guided vertices. The distillation procedure enables our feature extractor to own the ability to maintain insensitivity under vastly differentiated data distribution. Finally, we leverage an MLP decoder to predict the joint positions $\hat{\mathbf{J}}_{p}$. A combined loss $\mathcal{L}_{prior}$ consisting of $\mathcal{L}_{mse}(\hat{\mathbf{J}}_{p})$ and $\mathcal{L}_{pc}$ is utilized to train the network, which is formulated as below
\begin{equation}
    \begin{aligned}
    \mathcal{L}_{mse}(\hat{\mathbf{J}}_{p})= \sum_t \parallel \hat{\mathbf{J}}_{p}(t)-\widetilde{\mathbf{J}}^{GT}(t) \parallel_2^2,
    \end{aligned}
\label{equ:loss_Body}
\end{equation}

\begin{equation}
    \begin{aligned}
    \mathcal{L}_{pc}= \sum_t F_v(t) \log(\frac{F_v(t)}{F_p(t)}),
    \end{aligned}
\label{equ:loss_PC}
\end{equation}

\begin{equation}
    \begin{aligned}
    \mathcal{L}_{prior} = \lambda_1\mathcal{L}_{mse}(\hat{\mathbf{J}}_{p}) + \lambda_2\mathcal{L}_{PC},
    \end{aligned}
\end{equation}
where $\lambda_1$ and $\lambda_2$ are hyper-parameters, and we set $\lambda_1=1$ and $\lambda_2=10^3$ in our experiments. During inference, the VAD process is not required.

\subsection{Consecutive Pose Optimizer}
\label{sec:pose_optimizer}

\begin{figure}[ht]
	\centering
	\includegraphics[width=\linewidth]{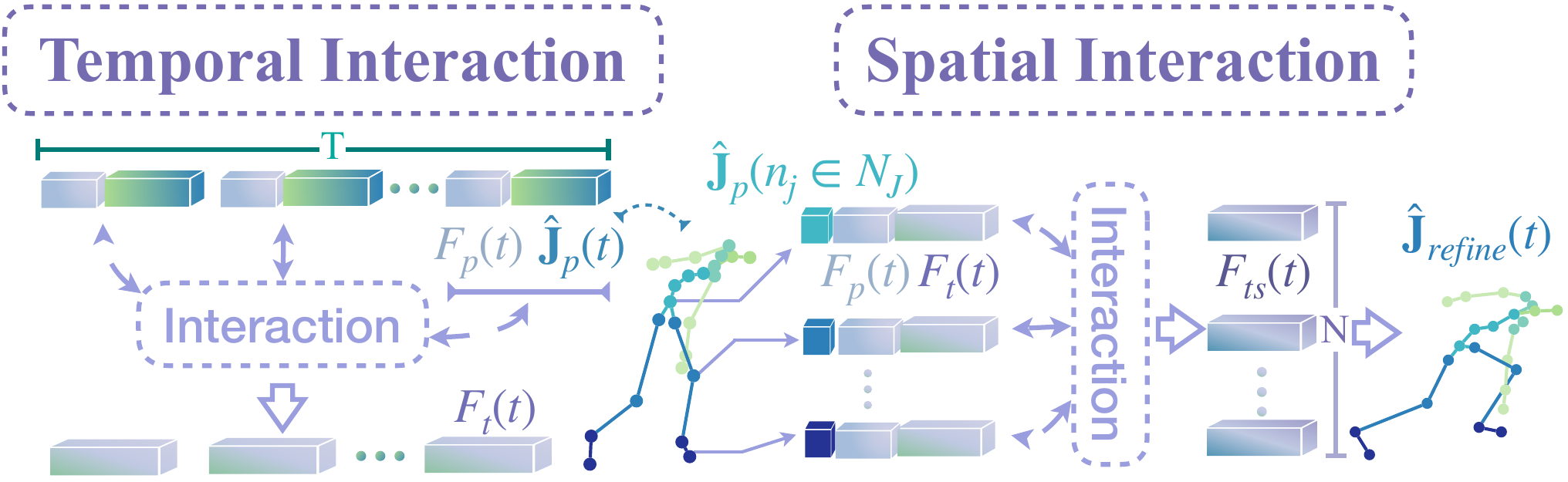}
	\caption{The detailed feature interaction mechanism in CPO. 
 The same network architecture is applied in both consecutive pose optimizer and multi-head solver(pose and shape) except the decoder. Here we take the consecutive pose optimizer as the reference.
 }
	\label{fig:CPO}
	\vspace{-2ex}
\end{figure}

We have already obtained the joint positions of human poses from the point-based body tracker. Considering that human motions are coherent at time sequence and different joints of the human body usually execute the action with relative dynamic constraints, we propose a \textbf{Consecutive Pose Optimizer (CPO)} (Fig.~\ref{fig:CPO}) to refine the body skeleton using consecutive joint-wise geometry features and relationship features in temporal and spatial spaces, which can further reduce the effect of incomplete and noised point clouds. 
Specifically, we utilize the concatenation of point-wise feature $F_p(t)\in \mathbb{R}^{k}$ and the predicted joint positions $\hat{\mathbf{J}}_{p}(t)$ as the initial joint-wise feature input. To capture the motion consistency in sequence, we use linear transformations to generate $Q(t)$, $K(t)$, and $V(t)$ in each frame and conduct temporal interaction to learn the motion-consistent feature $F_t(t)\in \mathbb{R}^{k_2}$ for each joint, where $k_2 = 256$. This temporal interaction process guides the estimation of more reasonable continuous human motions, especially for occluded situations.
Then, we use the dynamic and geometric constraints among joints to further enhance the joint feature via spatial feature interaction. The input $F_j(n_j\in N_J)\in \mathbb{R}^{k+k_2+3}$ consists of the point-wise feature $F_p(t)\in \mathbb{R}^{k}$, temporal interaction feature $F_t(t)\in \mathbb{R}^{k_2}$, and each joint feature $\hat{\mathbf{J}}_{p}(n_j\in N_J) \in \hat{\mathbf{J}}_{p}(t)$. We generate $Q(n_j)$, $K(n_j)$ and $V(n_j)$ with linear mapping for each joint and conduct the spatial joint-to-joint interaction to get the enhanced feature $F_{ts}(t)\in \mathbb{R}^{k_3}$, where $k_3=512$. The feature interaction matrix can be formulated as:

\begin{equation}
    \begin{aligned}
    \mathcal{F}_{interaction}=\operatorname{softmax}\left(\frac{Q K^T}{\sqrt{d_k}}\right) V.
    \end{aligned}
\label{eq:interaction}
\end{equation}
Finally, we regress the refined joint positions $\hat{\mathbf{J}}_{refine}(t)$ from the enhanced feature and the loss function is
\begin{equation}
    \begin{aligned}
    \mathcal{L}_{mse}(\hat{\mathbf{J}}_{refine})= \sum_t \parallel \hat{\mathbf{J}}_{refine}(t)-\widetilde{\mathbf{J}}^{GT}(t) \parallel_2^2 .
    \end{aligned}
\label{equ:loss_RefineJ}
\end{equation}

\subsection{Multi-head SMPL Solver}
\label{sec:IK}
In the last stage, we propose an attention-based multi-head solver to regress the SMPL~\cite{SMPL2015} parameters $\hat{\mathbf{\theta}}(t)$, $\hat{\mathbf{\beta}}$, $\hat{Tr}(t)$ from refined joint positions and the input point cloud. 
Because the pose and the shape reflect local geometry of human body, they can be determined by root-relative joint features obtained in last stage. We utilize the same network structure as CPO as the pose solver and shape solver to get $\hat{\mathbf{\theta}}(t)$ and $\hat{\mathbf{\beta}}$.
However, the global translation could be obtained only from the root-relative local geometry features. Previous methods~\cite{li2022lidarcap,ren2023lidar} usually take the average position of the body point cloud as the global location or directly regress the global translation. However, due to the interference of occlusion and noise, their predicted results are unstable in consecutive frames. In contrast, we simplify the task of predicting global translation to predict the bias between the average position of point cloud and the real 3D location. Thus, we propose a \textbf{Skeleton-aware Translation Solver} underpinned by a cross-attention architecture, which intelligently integrates skeletal and original point cloud data to get more accurate translation estimation.
We employ point cloud $\mathbf{P}(t)$ and refined root-relative joint positions $\hat{\mathbf{J}}_{refine}(t)$ as the input, utilizing the cross-attention to match the geometric information of joints with the point cloud. We generate the $Q(t)$ from refined joint positions and $K(t)$, $V(t)$ from point cloud. The feature interaction matrix can be formulated as Eq.~\ref{eq:interaction}. The decoder outputs the bias, which can be added to the average location $\mathbf{loc}(t)$ of raw point cloud to get the global translation $\hat{Tr}(t)$.
Finally, we use SMPL model to generate the human skeleton joint positions and mesh vertex positions as below.
\begin{equation}
    \begin{aligned}
    \hat{\mathbf{J}}_{smpl}(t),\hat{\mathbf{V}}_{smpl}(t) = \operatorname{SMPL}(\hat{\mathbf{\theta}}(t),\hat{\mathbf{\beta}},\hat{Tr}(t)).
    \end{aligned}
\label{equ:smpl}
\end{equation}
The loss function for the multi-head solver is formulated as:
\begin{equation}
    \begin{aligned}
    \mathcal{L}_{solver} = &\lambda_3\mathcal{L}_{mse}(\hat{\mathbf{J}}_{smpl})+\lambda_4\mathcal{L}_{mse}(\hat{\mathbf{V}}_{smpl})\\+&\lambda_5\mathcal{L}_{mse}(\hat{\mathbf{\theta}}(t))+\lambda_6\mathcal{L}_{mse}(\hat{\mathbf{\beta}}\\+&\lambda_7\mathcal{L}_{mse}(\hat{Tr}(t))+\lambda_8\mathcal{L}_{SUCD},
    \end{aligned}
\label{equ:solver_loss}
\end{equation}
where $\lambda_3$, $\lambda_4$, $\lambda_5$, $\lambda_6$, $\lambda_7$ are hyper-parameters with $\lambda_3 = \frac{100}{N_j}$, $\lambda_4 = \frac{100}{N_v}$, $\lambda_5 = \frac{1}{5}$, $\lambda_6 = 1$, $\lambda_7 = 1$ and $\lambda_8 = 10^3$.

Because the raw point cloud contains the real pose, shape, and global translation information, it can be taken as an extra supervision which is ignored by previous methods. In particular, we introduce a novel scene-level unidirectional Chamfer distance (SUCD) loss by calculating the unidirectional Chamfer distance from the raw point cloud to the predicted mesh vertices. It provides a comprehensive evaluation for all predicted SMPL parameters, denoted as

\begin{equation}
\begin{aligned}
\mathcal{L}_{SUCD}=\sum_t \frac{1}{|\mathbf{P}(t)|}\sum_{x \in \mathbf{P}(t)}\min_{y \in \hat{\mathbf{V}}_{smpl}(t)}\vert x-y \vert_{2}^2,
\end{aligned}
\label{equ:cdloss}
\end{equation}

%% file: sections/Dataset.tex
\begin{table*}
\centering
\caption{Comparison with public human motion datasets from four different aspects. ``Capture distance" means the maximum distance between performer and capture device, which is approximately calculated with the data published. ``Multi-person" indicates the capture scenes involve multiple persons. ``HOI" denotes the human-object interaction scenarios.}
\vspace{-2ex}
\label{tab:dataset}
\resizebox{\linewidth}{!}{
\begin{tabular}{c|cc|ccc|cccc|ccc}
    \toprule[1.2pt]
     \multirow{2}{*}{Dataset} & \multicolumn{2}{c|}{Statistics}& \multicolumn{3}{c|}{Scenarios}&\multicolumn{4}{c|}{Data}&\multicolumn{3}{c}{SMPL annotation}\\
    \cmidrule(r){2-3}
    \cmidrule(r){4-6}
    \cmidrule(r){7-10}
    \cmidrule(r){11-13}
      & Frame & Capture distance(m) & Multi-person & In the wild & HOI & Point cloud & IMU & Image & Multi-view & Pose & Shape & Translation  \\
    \midrule[1.2pt]
    AMASS~\cite{AMASS_ICCV2019} & 16M & 3.42 &\textcolor{red}{\XSolidBrush}  & \textcolor{red}{\XSolidBrush} & \textcolor{red}{\XSolidBrush} & \textcolor{red}{\XSolidBrush}& \textcolor{red}{\XSolidBrush}& \textcolor{green}{\Checkmark} & \textcolor{green}{\Checkmark}&\textcolor{green}{\Checkmark} &\textcolor{green}{\Checkmark} &\textcolor{green}{\Checkmark}\\
    HuMMan~\cite{cai2022humman} & 60M & 3.00 & \textcolor{red}{\XSolidBrush}& \textcolor{red}{\XSolidBrush}& \textcolor{red}{\XSolidBrush} &\textcolor{green}{\Checkmark}  &\textcolor{red}{\XSolidBrush} &\textcolor{green}{\Checkmark}&\textcolor{green}{\Checkmark}&\textcolor{green}{\Checkmark} &\textcolor{green}{\Checkmark} &\textcolor{green}{\Checkmark}\\
    SURREAL~\cite{varol2017learning}&6M&N/A&\textcolor{green}{\Checkmark}&\textcolor{red}{\XSolidBrush} &\textcolor{red}{\XSolidBrush} &\textcolor{red}{\XSolidBrush} &\textcolor{red}{\XSolidBrush} & \textcolor{green}{\Checkmark}&\textcolor{red}{\XSolidBrush}&\textcolor{green}{\Checkmark}&\textcolor{green}{\Checkmark}&\textcolor{green}{\Checkmark}\\
    AIST++~\cite{li2021learn} & 10M & 4.23 &\textcolor{red}{\XSolidBrush} & \textcolor{red}{\XSolidBrush} & \textcolor{red}{\XSolidBrush} & \textcolor{red}{\XSolidBrush} & \textcolor{red}{\XSolidBrush} & \textcolor{green}{\Checkmark} & \textcolor{green}{\Checkmark}&\textcolor{green}{\Checkmark}&\textcolor{red}{\XSolidBrush}&\textcolor{green}{\Checkmark} \\
    3DPW~\cite{von2018recovering} 
  & 51k & N/A &\textcolor{green}{\Checkmark} &\textcolor{green}{\Checkmark} & \textcolor{green}{\Checkmark} & \textcolor{red}{\XSolidBrush}& \textcolor{green}{\Checkmark}&\textcolor{green}{\Checkmark} & \textcolor{red}{\XSolidBrush} & \textcolor{green}{\Checkmark}&\textcolor{green}{\Checkmark} &\textcolor{green}{\Checkmark}\\
    LiDARHuman26M~\cite{li2022lidarcap} 
    & 184k & 28.05 &\textcolor{red}{\XSolidBrush} & \textcolor{green}{\Checkmark} & \textcolor{red}{\XSolidBrush} &\textcolor{green}{\Checkmark} &\textcolor{green}{\Checkmark} &\textcolor{green}{\Checkmark} & \textcolor{red}{\XSolidBrush}&\textcolor{green}{\Checkmark} &\textcolor{red}{\XSolidBrush} &\textcolor{red}{\XSolidBrush}\\
    LIPD~\cite{ren2023lidar} & 62k & 30.04  &\textcolor{red}{\XSolidBrush} &\textcolor{green}{\Checkmark} & \textcolor{red}{\XSolidBrush} & \textcolor{green}{\Checkmark}& \textcolor{green}{\Checkmark}&\textcolor{green}{\Checkmark} & \textcolor{red}{\XSolidBrush}&\textcolor{green}{\Checkmark}&\textcolor{red}{\XSolidBrush} &\textcolor{red}{\XSolidBrush}\\
    MOVIN~\cite{jang2023movin} &161k&N/A&\textcolor{red}{\XSolidBrush}&\textcolor{red}{\XSolidBrush}&\textcolor{red}{\XSolidBrush}&\textcolor{green}{\Checkmark}&\textcolor{red}{\XSolidBrush}&\textcolor{green}{\Checkmark}&\textcolor{green}{\Checkmark}&\textcolor{green}{\Checkmark}&\textcolor{red}{\XSolidBrush}&\textcolor{green}{\Checkmark}\\
    Sloper4D~\cite{dai2023sloper4d} &100k& N/A &\textcolor{red}{\XSolidBrush}&\textcolor{green}{\Checkmark} & \textcolor{red}{\XSolidBrush}&\textcolor{green}{\Checkmark} & \textcolor{green}{\Checkmark}& \textcolor{green}{\Checkmark}& \textcolor{red}{\XSolidBrush}&\textcolor{green}{\Checkmark} & \textcolor{green}{\Checkmark}&\textcolor{green}{\Checkmark}\\
    CIMI4D~\cite{yan2023cimi4d} & 180k & 16.61 & \textcolor{red}{\XSolidBrush}&\textcolor{green}{\Checkmark} &\textcolor{red}{\XSolidBrush} &\textcolor{green}{\Checkmark}&\textcolor{green}{\Checkmark} &\textcolor{green}{\Checkmark}& \textcolor{red}{\XSolidBrush} &\textcolor{green}{\Checkmark} & \textcolor{green}{\Checkmark}&\textcolor{green}{\Checkmark}\\
    \midrule[0.1pt]
    \textbf{FreeMotion} & 578k & 39.85 & \textcolor{green}{\Checkmark} & \textcolor{green}{\Checkmark} & \textcolor{green}{\Checkmark} & \textcolor{green}{\Checkmark} & \textcolor{green}{\Checkmark} & \textcolor{green}{\Checkmark} & \textcolor{green}{\Checkmark}& \textcolor{green}{\Checkmark}& \textcolor{green}{\Checkmark}& \textcolor{green}{\Checkmark} \\
    \bottomrule
\end{tabular}}
\vspace{-2ex}
\end{table*}

\section{FreeMotion Dataset}

Previous LiDAR-related human motion datasets typically involve a single performer carrying out common actions with incomplete SMPL parameters, which have limitations in evaluating the generalization capability and robustness of HPS methods when being applied in daily-life complex scenarios. To facilitate the research of high-quality human motion capture in large-scale free environment, we provide FreeMotion, the first motion dataset with multi-view and multi-modal visual data with full-SMPL annotations, captured in diverse real-life scenarios with natural occlusions and noise. It contains 578,775 frames of data and annotations and contains $1\sim7$ performers in each scene. 

	\vspace{-2ex}
\begin{figure}[ht]
	\centering
	\includegraphics[width=\linewidth]{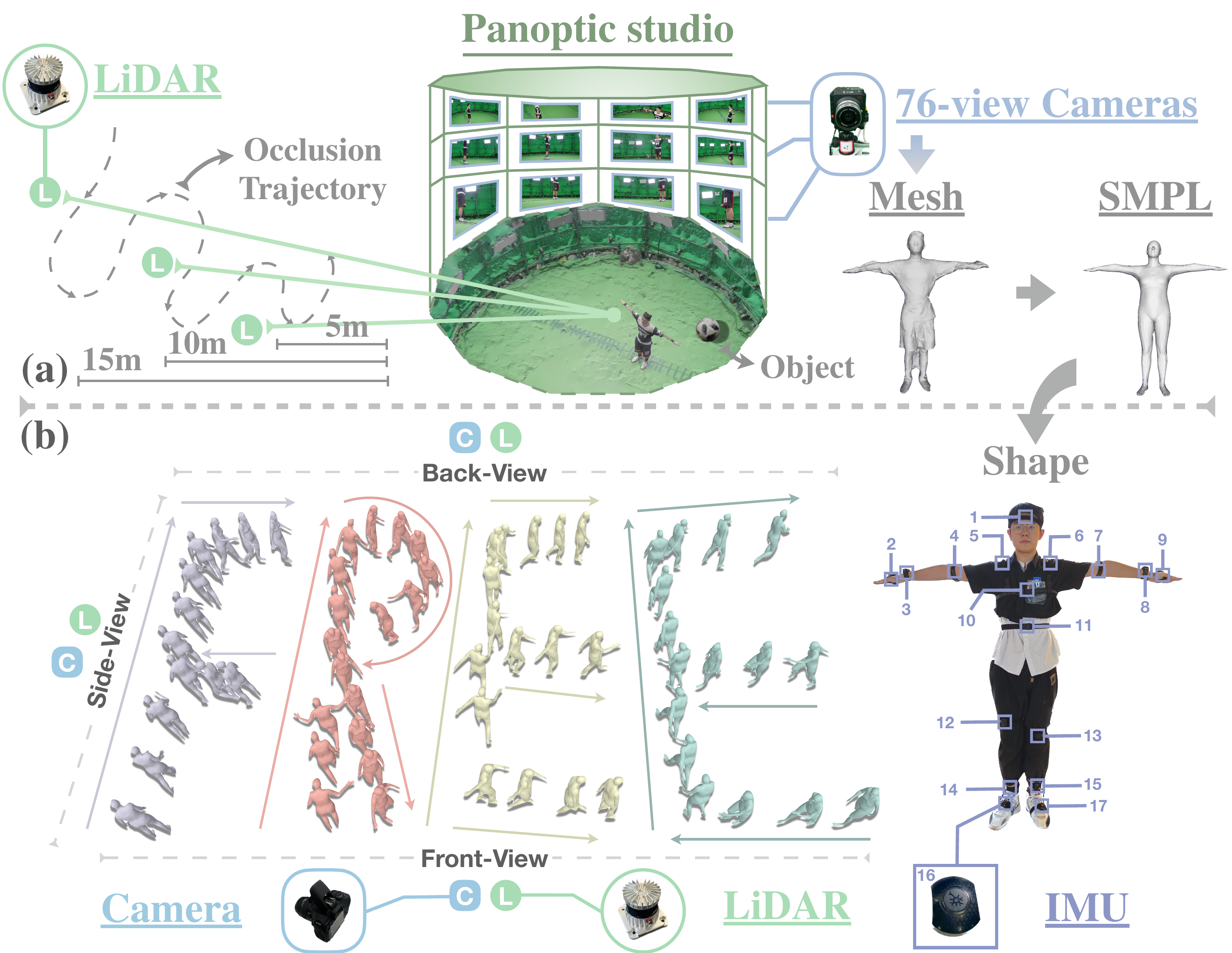}
	\caption{The capture systems of FreeMotion. In (a), we use a dense-camera capture system with LiDARs for accurate pose and shape capture. In (b), we set LiDARs and cameras at three views to capture human motions in large-scale multi-person scenes.}
	\label{fig:dataset}
	\vspace{-2ex}
\end{figure}

\subsection{Data Acquisition}\label{subsec2}
Considering that the indoor multi-camera panoptic studio can provide high-precision full SMPL parameter annotations (pose, shape and global translation) and outdoor scenes are large-scale and suitable for real applications, we have two capture systems as shown in Fig.~\ref{fig:dataset}. For the first one, we set up a 76-Z-CAM system to obtain SMPL ground truth and three OUSTER-1 LiDARs at varied distances to get LiDAR data. Notably, we arrange other performers outside the studio to simulate occlusions in real-life scenarios. For the second one, we built three sets of LiDAR-camera capture devices, including a 128-beam OUSTER-1 LiDAR and a monocular Canon camera for each, in different locations to capture multi-view and multi-range visual data, and provide the global translation ground truth. The performer is equipped with a full set of Notiom equipment (17 IMUs) to obtain the pose ground truth. Particularly, the shape parameters of outdoor performers are captured in panoptic studio in advance. The capture frequencies for the LiDAR, Z-CAM, Canon camera, and IMU are set at 10Hz, 25Hz, 60Hz, and 60Hz, respectively. All the data are calibrated and synchronized.

\subsection{Dataset Characteristics}
The detailed comparison with existing public datasets is presented in Tab.~\ref{tab:dataset}. FreeMotion has several distinctive characteristics and we summarize three main highlights below.

\textbf{Free Capture Scenes.}
Diverging from previous datasets focused on single-person HPS, FreeMotion is captured in real unconstrained environments, which involves diverse capture scales, multi-person activities (sports, dancing shows, fitness exercises, etc.), and human-object interaction scenarios. The large-scale human trajectories, occlusions, and noise in our data all bring challenges for precise human global pose and shape estimation, thereby promoting the envelope of HPS technology for real-life applications.

\textbf{Diverse Data Modalities and Views.}
FreeMotion offers multi-view and multi-modal capture data, including LiDAR point clouds, RGB images, and IMU measurements, providing rich resources for the exploration of single-modal, multi-modal, single-view, multi-view HPS solutions.



\textbf{Complete Scene-level SMPL Annotations.} 
Existing LiDAR-based motion datasets usually provide pose annotations using dense IMUs and lack annotations for accurate human shape and global translation. 
FreeMotion remedies this by providing full SMPL parameters annotations(pose, shape, translation), as shown in Fig.~\ref{fig:dataset}. We capture a variety of natural human motions across long distances, involving 20 individuals with varying body types engaging in 40 types of actions. Details are in appendix. Accurate and complete annotations in rich scenarios can comprehensive evaluation for algorithms and benefit many downstream applications. 

\subsection{Data Extension}
\label{sec:dataex}
To enrich the dataset with various poses and shapes for pretraining, we follow LIP~\cite{ren2023lidar} to create synthetic point clouds from SURREAL~\cite{varol2017learning}, AIST++~\cite{li2021learn}, and portions of AMASS~\cite{AMASS_ICCV2019}, including ACCAD and BMLMovi. We simulate occlusions by randomly cropping body parts of point clouds. It consists of $2,378$k frames and $3,118$ body shapes. Note that statistics in Tab.~\ref{tab:dataset} do not include the synthetic data. \textit{Detailed process is shown in appendix.}


\subsection{Privacy Preservation}
We adhere to privacy guidelines in our research. The LiDAR point clouds naturally protect privacy by omitting texture or facial details. Additionally, we mask faces in RGB images to uphold ethical standards in our dataset.


%% file: sections/Experiments.tex
\section{Experiments}\label{sec5}

\begin{table*}[ht!]\normalsize
\centering
\caption{Comparison with state-of-the-art methods on various datasets. Lower values represent better performance for all metrics.}
\vspace{-2ex}
\resizebox{\linewidth}{!}{
\begin{tabular}{cccccc|ccccc|ccccc}
\toprule
\multirow{2}{*}{} & \multicolumn{5}{c|}{SURREAL~\cite{varol2017learning}}&\multicolumn{5}{c|}{Sloper4D~\cite{dai2023sloper4d}}&\multicolumn{5}{c}{FreeMotion}\\
\cmidrule(r){2-6}
\cmidrule(r){7-11}
\cmidrule(r){12-16}
& J/V Err(P)$\downarrow$&J/V Err(PS)$\downarrow$&J/V Err(PST)$\downarrow$&Ang Err$\downarrow$&SUCD$\downarrow$&J/V Err(P)$\downarrow$&J/V Err(PS)$\downarrow$& J/V Err(PST)$\downarrow$& Ang Err$\downarrow$ & SUCD$\downarrow$ & J/V Err(P)$\downarrow$& J/V Err(PS)$\downarrow$& J/V Err(PST)$\downarrow$& Ang Err$\downarrow$& SUCD$\downarrow$ \\
\midrule
LiDARCap~\cite{li2022lidarcap}&42.82/54.05&51.05/62.42&118.51/123.84&9.90&3.02
&67.40/80.08&71.99/86.58&179.33/185.39&15.92&4.54
&87.58/105.97&88.98/107.64&186.55/196.06&16.79&5.00\\
LIP~\cite{ren2023lidar}&31.72/42.40&32.71/43.02&45.22/53.18&12.17&0.95
&60.11/74.90&62.03/77.26&94.81/106.34&19.95&2.27
&81.13/98.65&81.99/99.58&129.88/141.77&18.76&4.24\\
MOVIN~\cite{jang2023movin}&97.34/120.49&103.37/125.64&-&26.98&-
&123.80/146.25&126.19/148.69&45559.25/45564.41&32.12&3311762.48
&109.62/128.66&113.47/132.25&2853.87/2863.89&27.34&9252.86\\
\midrule
\textbf{LiveHPS}&\textbf{23.99/30.81}&\textbf{24.75/31.81}&\textbf{34.45/40.14}&\textbf{9.49}&\textbf{0.67} 
&\textbf{46.22/56.72}&\textbf{48.28/59.02}&\textbf{77.73/85.83}&\textbf{12.77}&\textbf{1.67}
&\textbf{68.88/83.20}&\textbf{69.43/83.90}&\textbf{119.27/128.61}&\textbf{15.79}&\textbf{2.85}\\
\bottomrule
\end{tabular}}
\label{tab:compare}
\vspace{-2ex}
\end{table*}

\begin{table*}[ht!]\scriptsize
\centering
\caption{The cross-dataset evaluation on various datasets. We use applicable metrics for each dataset according to its annotations.}
\vspace{-2ex}
\resizebox{\linewidth}{!}{
\begin{tabular}{cccc|ccc|cccc|c|c}
\toprule
\multirow{2}{*}{} & \multicolumn{3}{c|}{LiDARHuman26M~\cite{li2022lidarcap}}&\multicolumn{3}{c|}{LIPD~\cite{ren2023lidar}}&\multicolumn{4}{c|}{CIMI4D~\cite{yan2023cimi4d}}&SemanticKITTI~\cite{behley2019semantickitti}&HuCenLife~\cite{xu2023human}\\
\cmidrule(r){2-4}
\cmidrule(r){5-7}
\cmidrule(r){8-11}
\cmidrule(r){12-13}
& J/V Err(P)$\downarrow$&Ang Err$\downarrow$&SUCD$\downarrow$&J/V Err(P)$\downarrow$& Ang Err$\downarrow$ & SUCD$\downarrow$ & J/V Err(P)$\downarrow$& J/V Err(PS)$\downarrow$& Ang Err$\downarrow$& SUCD$\downarrow$&SUCD$\downarrow$&SUCD$\downarrow$ \\
\midrule
LiDARCap~\cite{li2022lidarcap}& 123.09/151.55&26.41&5.81&97.41/119.89&18.48&4.30&205.24/253.58&205.51/255.58&32.68&14.40&10.07&6.01\\
LIP~\cite{ren2023lidar}&103.48/124.18&24.14&3.93&83.38/102.25&18.73&2.81&162.28/205.25&166.38/211.10&33.03&8.47&9.93&5.06\\
MOVIN~\cite{jang2023movin}& 104.89/127.32 &32.56&188906.16&101.78/121.67&28.82&66400.65&178.48/214.16&182.29/218.07& 42.41& 39681.84&1647852.73&58655.42\\
\midrule
\textbf{LiveHPS}&\textbf{101.33/121.74}&\textbf{23.58}&\textbf{2.67}&\textbf{78.63/97.45}&\textbf{18.36}&\textbf{1.98}&\textbf{142.00/181.21}&\textbf{149.42/190.60}&\textbf{32.17}&\textbf{4.26}&\textbf{7.28}&\textbf{3.14}\\
\bottomrule
\end{tabular}}
\label{tab:compare_cross}
\vspace{-4ex}
\end{table*}


\begin{figure}[ht!]
	\centering
	\includegraphics[width=\linewidth]{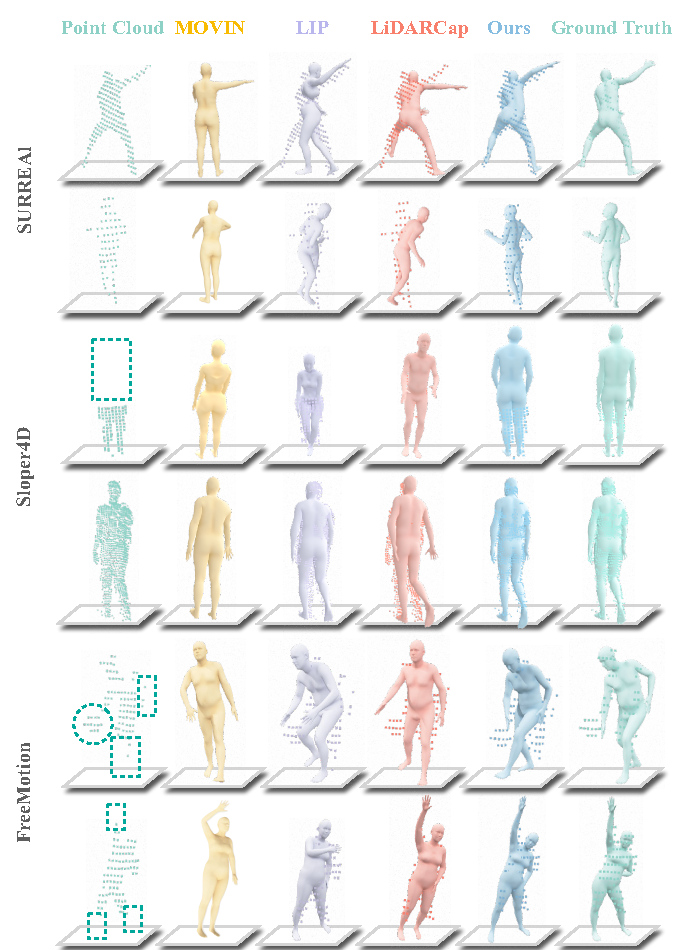}
        \vspace{-4ex}
	\caption{Qualitative comparisons. The point cloud matches the result better, representing more accurate estimation for pose, shape, and translation. Point cloud is far from results of MOVIN.}
	\label{fig:compare}
	\vspace{-3ex}
\end{figure}

\begin{figure}[ht!]
	\centering
	\includegraphics[width=\linewidth]{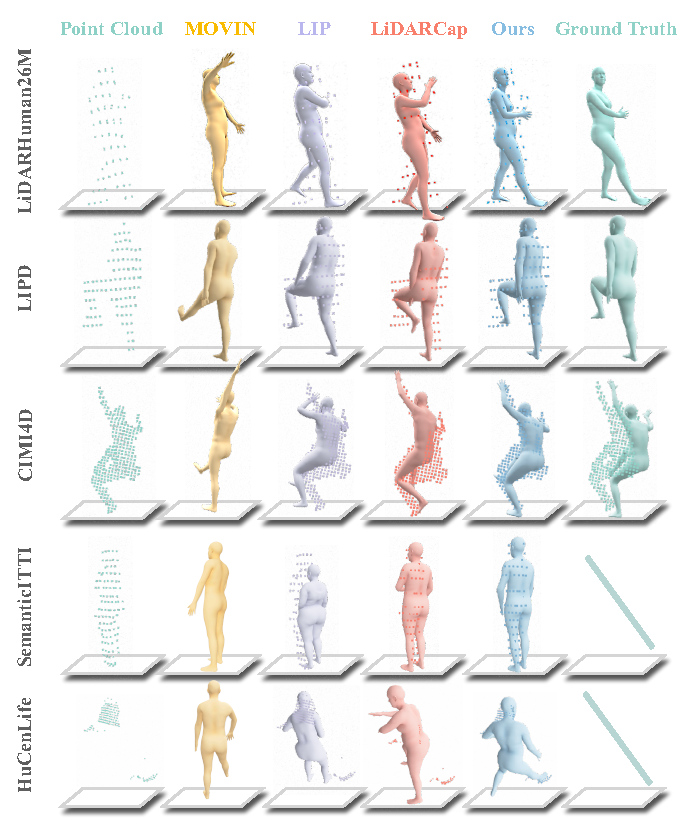}
	\caption{Qualitative comparisons in cross-dataset evaluation. 
 SemanticKITTI and HucenLife do not provide SMPL annotations.}
	\label{fig:compare_2}
	\vspace{-5ex}
\end{figure}

In this section, we compare our method with current SOTA methods on FreeMotion and various public datasets qualitatively and quantitatively, demonstrating our method's superiority and generalization capability. We also present detailed ablation studies for our network's modules to validate their effectiveness. Our evaluation metrics include 1) J/V Err(P/PS/PST) $\downarrow$: mean per joint/vertex position error in millimeters, where we generate joint/vertex from SMPL model by Pose/Pose-Shape/Pose-Shape-Translation parameters; 2) Ang Err $\downarrow$: mean per global joint rotation error in degrees to evaluate local pose; 3) SUCD $\downarrow$: scene-level unidirectional Chamfer distance in millimeters.


\begin{table*}[t]\normalsize
\centering
\caption{Ablation studies for our network modules. We also evaluate the internal details of each module.}
\vspace{-2ex}
\resizebox{\linewidth}{!}{
\begin{tabular}{c|cc|cc|cc|ccc|c}
    \toprule
    \multirow{2}{*}{} & \multicolumn{2}{c|}{Network Module} &  \multicolumn{2}{c|}{Consecutive Pose Optimizer} & \multicolumn{2}{c|}{Multi-head SMPL(Pose and Shape) Solver}&\multicolumn{3}{c|}{Skeleton-aware Translation Solver}&\multirow{2}{*}{Ours}\\ 
    \cmidrule(r){2-3}
    \cmidrule(r){4-5}
    \cmidrule(r){6-7}
    \cmidrule(r){8-10}
    & w/o VAD & w/o CPO & w/o Temporal & w/o Spatial & ST-GCN & GRU& Average & MOVIN & LIP & \\
    \midrule
    J/V Err(PST) & 129.19/140.42& 127.44/140.87 &121.93/135.56&120.20/129.51
    &124.37/135.38&120.63/130.83&177.66/184.57&1296.48/1310.95 &165.04/172.36 &\textbf{119.27/128.61} \\
    Ang Err &16.95&25.20& 27.58&18.09&19.40&18.34&-&-&-&\textbf{15.79}\\
    SUCD & 3.17&4.08& 3.51&2.95&3.28&3.08 &4.25& 8569.68&3.07 &\textbf{2.85}\\
    \bottomrule
\end{tabular}}
\vspace{-4ex}
\label{tab:ablation-s}
\end{table*}

\subsection{Implementation Details}
We build our network on PyTorch 1.8.1 and CUDA 11.1, trained over 200 epochs with batch size of 32 and sequence length of 32, using an initial learning rate of $10^{-3}$, and AdamW optimizer with weight decay of $10^{-4}$. The process was run on a server equipped with an Intel(R) Xeon(R) E5-2678 CPU and 8 NVIDIA RTX3090 GPUs. For training, we used clustered and manually annotated human point cloud sequences from raw data, while for testing, we employ sequential point clouds of human instances processed by a pre-trained segmentation model~\cite{vu2022softgroup}. As for the dataset splitting, we take training set of FreeMotion, Sloper4D, and synthetic dataset including training set of SURREAL, AIST++, ACCAD, and BMLMovi for training. 

\subsection{Comparison}
We evaluate LiveHPS against other state-of-the-art (SOTA) LiDAR-related methods~\cite{li2022lidarcap,ren2023lidar,jang2023movin} on FreeMotion and several public datasets~\cite{varol2017learning,dai2023sloper4d,li2022lidarcap,ren2023lidar,yan2023cimi4d,behley2019semantickitti,xu2023human} to demonstrate its superiority in capturing human global poses and shapes in large-scale free environment, even with severe occlusions and noise. 
Our LiveHPS achieves SOTA performance as shown in Tab.~\ref{tab:compare}. The J/V Err(P) and Ang Err only relate to the pose parameter estimation, we surpass LiDARCap~\cite{li2022lidarcap}, LIP~\cite{ren2023lidar} and MOVIN~\cite{jang2023movin} by an obvious margin.
For fair comparison, we only use the LiDAR branch of LIP. 
As the pioneer to fully estimate SMPL parameters (pose, shape, global translation) for LiDAR-based HPS, we develop a shape regression head with the same architecture of their pose regression head for fair comparison with other methods~\cite{li2022lidarcap, ren2023lidar, jang2023movin}, the translation prediction of LiDARCap is the average of point cloud.
Visual comparisons in Fig.~\ref{fig:compare} further highlight our method's superiority in global pose and shape estimations, yielding results that closely mirror ground truth. Other methods struggle in situations with occlusions and noise, as exemplified in challenging scenes from Sloper4D~\cite{dai2023sloper4d} and FreeMotion in Fig.~\ref{fig:compare}.
MOVIN~\cite{jang2023movin} estimate translation based on velocity regression, it is not applicable on synthetic data SURREAL without real trajectories. 
Our LiveHPS demonstrates robust performance against noise like carried objects, as demonstrated in FreeMotion's left case in Fig.~\ref{fig:compare}.

Tab.~\ref{tab:compare_cross} illustrates our cross-dataset evaluation to validate the generalization capability of LiveHPS by directly testing on other datasets. LiDARHuman26M~\cite{li2022lidarcap} and LIPD~\cite{ren2023lidar} only offer pose parameters. CIMI4D~\cite{yan2023cimi4d} provides pose, shape, and translation, but the translation is not that precise as shown in the third row of Fig.~\ref{fig:compare_2}. SemanticKITTI~\cite{behley2019semantickitti} and HuCenLife~\cite{xu2023human} are large-scale datasets for 3D perception, not providing SMPL annotations. 
Thanks to our VAD module's ability to harmonize diverse human point cloud distributions and CPO module's ability to model geometric and dynamic human features, our method can achieve SOTA performance on these cross-domain datasets, even in challenging cases with extreme occlusions, as Fig.~\ref{fig:compare_2} shows.

\subsection{Ablation Study}
We first validate the effectiveness of each module in LiveHPS. Then, we evaluate inner designs of each module to verify the effectiveness of detailed structures. 

Tab.~\ref{tab:ablation-s} shows the performance of our method with different network modules, demonstrating the necessity of our vertex-guided adaptive distillation (VAD) and consecutive pose optimizer (CPO) modules. 
We also illustrate ablation details of attention-based temporal and spatial feature enhancement in CPO, showing that the combination of temporal and spatial feature interaction performs best. We also conduct experiments to validate our attention-based multi-head SMPL solver. Our pose and shape solver, using the same network as CPO, outperforms ST-GCN from LiDARCap~\cite{li2022lidarcap} and GRU from LIP~\cite{ren2023lidar} by fully utilizing the global temporal context and local spatial relationship existing in consecutive body joints. 
For the translation solver, the average of point cloud can reflect the coarse translation but it is very unstable with the distribution of point cloud changes. Compared with global velocity estimation utilized in MOVIN~\cite{jang2023movin}, our skeleton-aware translation solver directly estimates translations without error accumulation. 
Moreover, unlike GRU-based pose-guided corrector in LIP~\cite{ren2023lidar} which overlooks relationship between the skeleton and point cloud, our approach performs better by considering the relationship and more spatial information.


\begin{table}[ht!]
\centering
\caption{More results on different lengths of input sequence and different point numbers on humans on FreeMotion dataset.}
\vspace{-2ex}
\resizebox{\linewidth}{!}{
\begin{tabular}{cccccc}
    \toprule
    Frames &1&4&8&16&32\\
    \midrule
    J/V Err(PST)$\downarrow$&142.88/155.58&130.73/141.10&126.23/135.60& 123.08/132.14& \textbf{119.27/128.61}\\
    Ang Err$\downarrow$&19.22&17.31&16.53&16.05&\textbf{15.79}\\
    SUCD$\downarrow$&5.22&3.03 &3.01& 3.02& \textbf{2.85} \\
    \midrule
    \midrule
    Points &$0\sim100$&$100\sim200$&$200\sim300$&$300\sim1000$&$>1000$\\
    \midrule
    J/V Err(PST)$\downarrow$&156.01/168.42&106.03/114.00&106.81/113.98&103.96/110.37&81.01/87.70\\
    Ang Err$\downarrow$&16.78&16.34&15.17&13.64&12.84\\
    SUCD$\downarrow$&4.54&2.31&2.25&2.52&2.63\\
    \bottomrule
\end{tabular}}
\label{tab:difT}
\vspace{-4ex}
\end{table}

\subsection{Generalization Capability Test}
\begin{figure}[ht]
	\centering
	\includegraphics[width=0.95\linewidth]{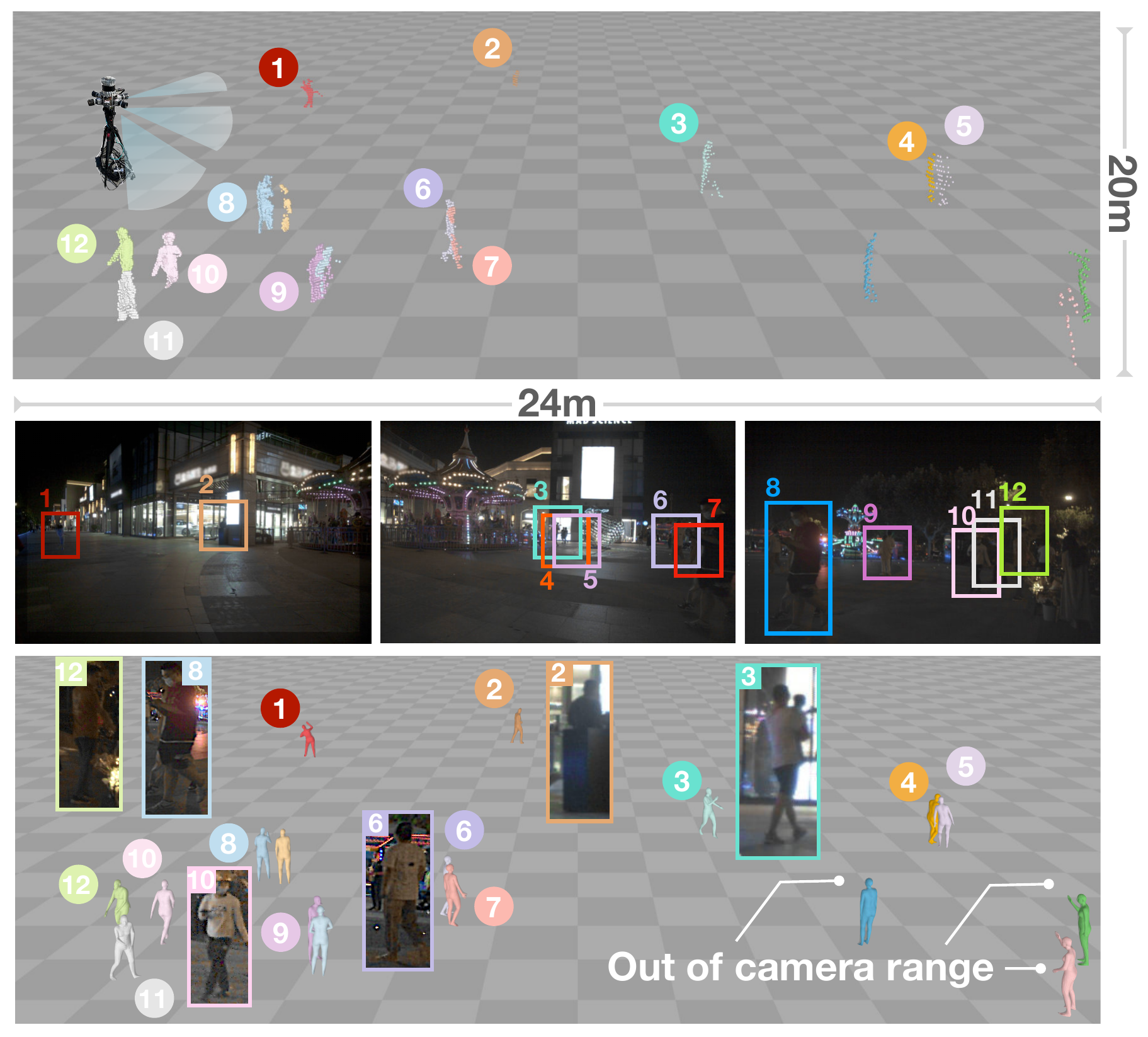}
 	\vspace{-2ex}
	\caption{Performance of LiveHPS on real-time-captured scenes. 
 }
	\label{fig:dark}
	\vspace{-4ex}
\end{figure}

We assess the generalization capability of LiveHPS across varying lengths of input point cloud sequences and across different point numbers on human body in each frame, as Tab.~\ref{tab:difT} shows. Our method performs better with increasing sequence length but maintains good accuracy even with short inputs. In addition, our method performs relatively robust even for the situation with 100 points on the human body, which means far distance (about 15 meters) to LiDAR or severe occlusion. Fig.~\ref{fig:teaser} and Fig.~\ref{fig:dark} show our method is practical for in-the-wild scenarios, capturing human motion in large-scale scenes day and night with real-time performance up to 45 fps. This strongly demonstrates the feasibility and superiority of our method in real-life applications. \textit{More application results are in appendix.}

\vspace{-1ex}

%% file: sections/Conclusion.tex
\section{Conclusion}
In this paper, we propose a novel single-LiDAR-based approach for predicting human pose, shape, and translations in large-scale free environment. To solve the occlusion and noise interference, we design a novel distillation mechanism and temporal-spatial feature interaction optimizer.
Importantly, we propose a huge multi-person human motion dataset, which is significant for future in-the-wild HPS research. Extensive experiments on diverse datasets demonstrate the robustness and effectiveness of our method.

\textbf{Limitations} When human is static in the large-scale scene for a long time, our model can not fully utilize the dynamic information in consecutive frames and cause the misjudged orientation of human global orientations, opposite to the ground-truth pose. 

%% file: sections/sup.tex
\clearpage
\setcounter{page}{1}
\maketitlesupplementary

\input{sup_sections/intro}    
\input{sup_sections/calibration}
\input{sup_sections/simulation}
\input{sup_sections/dataset}
\input{sup_sections/more_visul}
\input{sup_sections/more_results}

%% file: sup_sections/intro.tex
In the supplementary material, we first provide a detailed explanation of our data processing procedures. Our capture system for the newly collected dataset encompasses multiple sensors, such as LiDARs, cameras, and IMUs. In Section.~\ref{sec:cali}, we meticulously present the details of data processing. 
In order to enhance the dataset's diversity of actions and shapes for pretraining, we generate a large amount of synthetic data. Section.~\ref{sec:data_syn} elaborates on the methodology employed for generating synthetic point cloud data. 
Additionally, we emphasize that the FreeMotion dataset encompasses various shapes and actions with occlusions and human-object interactions. In Section.~\ref{Sec:dataset}, we showcase nearly twenty human scans and forty distinct action types in FreeMotion. Moreover, in Section.~\ref{sec:more_ex}, we show more experiment and qualitative evaluations of our ablation study.
Lastly, in Section.~\ref{Sec:more_r}, we present multi-frame results of LiveHPS in diverse scenarios, containing indoor, outdoor, and night scenes, to further demonstrate the robustness and effectiveness of our method for free HPS in arbitrary scenarios.

%% file: sup_sections/calibration.tex
\section{Details of Data Processing}
\label{sec:cali}

FreeMotion is collected in two capture systems, the first capture system is the indoor multi-camera panoptic studio and the second capture system is in various challenging large-scale scenarios with multiple types of sensors. In this section, we provide more details about our data acquisition, data pre-processing, multi-sensor synchronization and calibration. The whole FreeMotion dataset will be made publicly available. 

\subsection{Data Acquisition}
When we capture data in both capture systems, we divide the capture processing into three stages. In the first stage, the actor performs different kinds of actions alone without occlusion and noise. In the second stage, we arrange other persons to interact with main actor and perform specific actions depending on the scene characteristics, such as basketball court, meeting room, etc. The interactions bring real occlusions. In the last stage, the main actor interacts with some objects, such as the balls, package skateboard, etc. The objects bring real noise.

\subsection{Data Pre-Processing}
\textbf{Point Cloud.} The raw point clouds are captured from 128-beam OUSTER-1 LiDAR. The LiDAR features a 360° horizon field of view (FOV) $\times$ 45° vertical FOV. For the training dataset, we first record the point cloud of static background, and then set a threshold to delete the background points to get the point clouds of actors. For the processed point clouds, we use DBSCAN~\cite{ester1996density} cluster to get instance point cloud of each person. Then we employ the Hungarian matching algorithm for point cloud-based 3D tracking. Moreover, to promise high-quality annotations of our dataset, we review the segmentation point cloud sequence of each person. For the challenge and complex scene, such as the meeting room, which contains many extra objects, we use manually annotation for each point cloud sequence. 
For the testing dataset, we train the instance segmentation model~\cite{vu2022softgroup} in our training set and inference in our testing set while the tracking processing is the same by above methods.

\textbf{SMPL annotations.} For the first capture system, we use multi-camera method~\cite{easymocap} to generate the ground-truth SMPL parameters(Poses and translations). The predicted shape parameters is not accurate by this method. As the Figure~\ref{fig:diverse_shape} shows, we use the multi-camera method~\cite{rc} to reconstruct the human mesh and fit the human mesh to the SMPL model in template pose for more accurate shape parameters. For the second capture system, we use full set of Noitom~\cite{noitom} equipment(17 IMUs) to get the SMPL pose parameters and the shape parameters of performer are captured in the panoptic studio in advance. In outdoor capture scenes, we follow ~\cite{dai2023sloper4d} to fit the SMPL mesh to the point cloud for accurate translation parameters.

\subsection{Multi-sensor System Synchronization and Calibration} \label{sec:calibration}
\textbf{Synchronization}
The synchronization among various-view and modal sensors is accomplished by detecting the jumping peak shared across multiple devices. Thus, actors are asked to perform jumps before the capture. We subsequently manually identify the peaks in each capture device and use this timestamp as the start for time synchronization. 

\textbf{Calibration}
Our capture systems contain LiDARs, Cameras and IMUs. For the three LiDARs calibration, we select the static background point clouds in the same timestamp, and manually to do the point cloud registration. For multi-camera calibration, we use ~\cite{rc} to calibrate all cameras. The calibration of IMUs is the algorithm of Noitom~\cite{noitom}.
For calibration of each sensor, in the first capture system, we generate the SMPL human mesh from multi-camera data, the coordinate of human mesh is the same with the multi-camera coordinate, then we follow LIP~\cite{ren2023lidar} to utilize the ICP~\cite{besl1992method} method between the segmentation human point cloud and the human mesh vertices for LiDAR-camera calibration. In the second capture system, we utilize the Zhang's Camera Calibration Method~\cite{zhang2000flexible} for LiDAR-camera calibration, and the human SMPL mesh's coordinate is the same with the coordinate of IMUs. We use the above ICP method for LiDAR-IMU calibration.

%% file: sup_sections/simulation.tex
\section{Data Synthesis}
\label{sec:data_syn}
In order to ensure the diversity of training data, we synthesize point cloud data on public datasets~\cite{AMASS_ICCV2019,li2021learn,varol2017learning}, which do not contain point cloud as input. In this section, we will explain our detailed implementations of data synthesis.
LiDAR works in a time-of-flight way with simple principles of physics, which can be easily simulated with a small gap to real data. We generate simulated LiDAR point cloud by emitting regular lights from the LiDAR center according to its horizontal resolution and vertical resolution. The light can be reflected back when encountering obstacles, generating a point at the intersection on the surface of obstacles. We conduct the simulation according to the parameters of Ouster(OS1-128), which is also the device used in collecting FreeMotion. Its horizontal resolution is 2048 and its vertical resolution is 128 lines. Each emission direction is described by unit vector in spherical coordinate system $d = [\cos{\varphi}\sin{\theta},\cos{\varphi}\cos{\theta},\sin{\varphi}]$, where $\varphi$ represents the angle between the emission direction and the plane XY, $\theta$ indicates the azimuth, and $c = [0,0,2]$ is the LiDAR center. The intersection point $p = [p_x,p_y,p_z]$ is calculated by 
\begin{equation}
    \begin{aligned}
    p = c + d\frac{n^T(q-c)}{n^Td},
    \end{aligned}
\label{equ:point}
\end{equation}
where $n$ represents the normal vector of corresponding mesh and $q$ denotes any vertex point of the mesh.

For the process of calculating the intersection of LiDAR and mesh surface, there are mainly three steps. The first step is to calculate the intersection of light and triangular patch, and the second step is to judge whether the intersection of light and plane is inside the triangle. Due to occlusions, one light should only have the intersection with the first touched mesh of object. Finally, we filter the intersections to only keep the ones first occurred in the LiDAR view.

Since the above datasets focus on single-person scenarios, we randomly crop some body parts of the point clouds to simulate the occlusions in real scenes:
\begin{equation}
    \begin{aligned}
    index &= dis(p - o) > r,\\
    pc_{crop} &= pc[index],
    \end{aligned}
\label{equ:sup_point}
\end{equation}
where $o$ represents a random point position in point cloud $pc$ as the round dot, and $r$ denotes the radius of the crop area. The function $dis()$ means the distance between the two point positions. The $pc_{crop}$ is the synthetic point cloud with occlusion.

%% file: sup_sections/dataset.tex
\section{Details of Dataset}
\begin{figure*}[t]
	\centering
	\includegraphics[width=\linewidth]{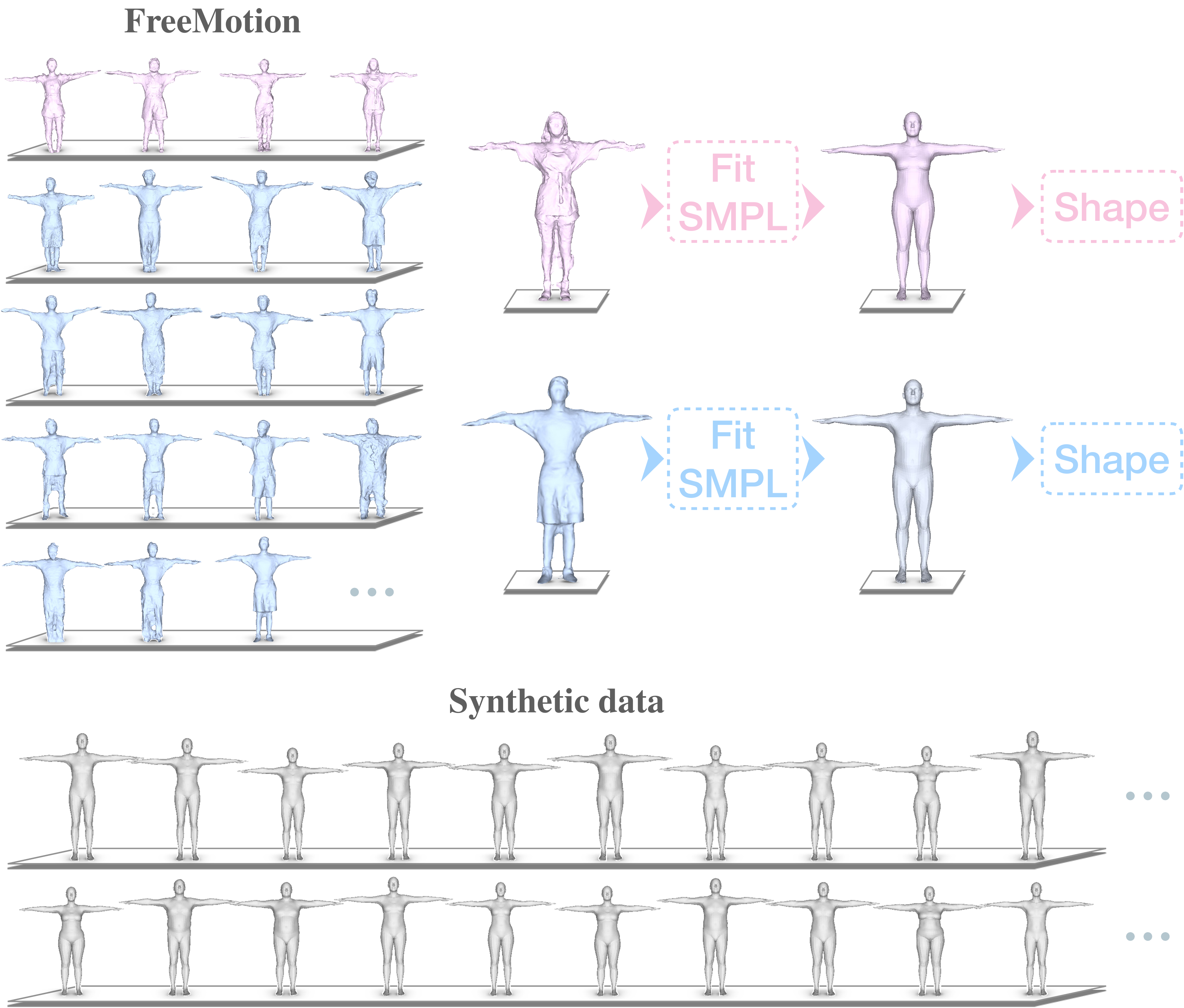}
	\caption{Diverse shapes of human in FreeMotion and synthetic dataset. We display the human reconstruction meshes and the process of generating shape parameters.}
	\label{fig:diverse_shape}
	\vspace{-3ex}
\end{figure*}

\begin{figure*}[t]
	\centering
	\includegraphics[width=\linewidth]{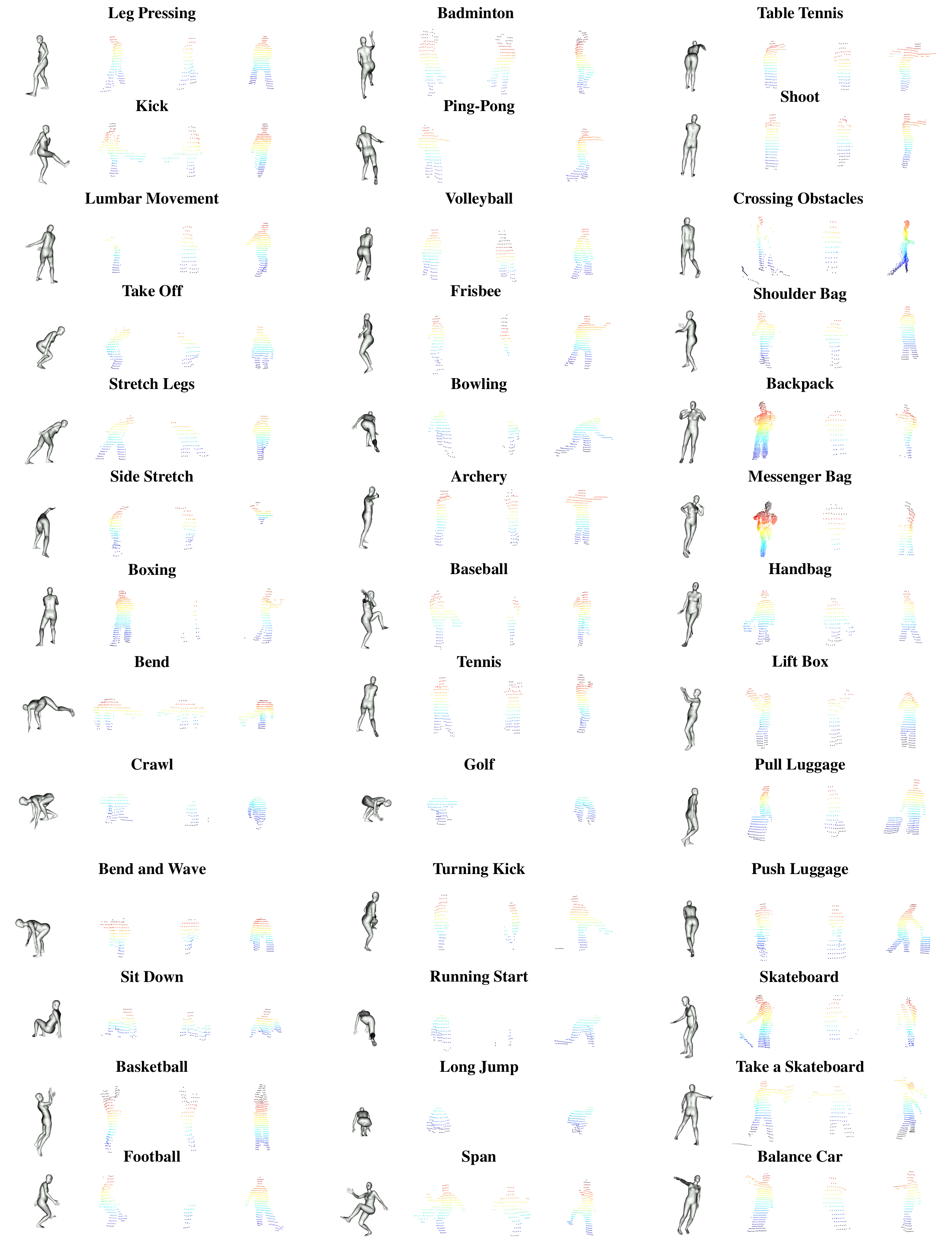}
	\caption{Different types of action in FreeMotion. We display the point cloud from three different views and corresponding ground-truth mesh for each type of action. The left point cloud is the main view, which is matching with the ground-truth mesh, the middle point cloud is the back view and the right point cloud is the side view.}
	\label{fig:sup_dataset}
	\vspace{-3ex}
\end{figure*}
\label{Sec:dataset}
We collect FreeMotion in diverse multi-person scenarios, including various sports venues as well as daily scenes. We calculate the types of action and the shapes of human covered in different scenarios. As the Figure.~\ref{fig:diverse_shape} shows, FreeMotion contains diverse human shapes, and we use large synthetic data to obtain more human with different shapes, such as SURREAL~\cite{varol2017learning}. As the Figure.~\ref{fig:sup_dataset} shows, FreeMotion comprises nearly forty distinct action types, providing point cloud data from three different views and corresponding ground-truth mesh.
The action types are arranged from top to bottom and from left to right, we begin by showing some common warm-up actions performed in daily routines, such as the leg pressing, kick or lumbar movement, etc. These fundamental motions are essential for sports activities and represent simple motions with minimal self-occlusion. However, they are still affected by external occlusion, as depicted in the left point cloud of the "Lumbar Movement" and the right point cloud of the "Side Stretch".
The more challenging daily motion type for point cloud is the squatting, which involves serious self-occlusion, such as the bend, crawling, and sitting down. Moreover, squatting motions are prone to lose information due to external occlusion, as shown in the middle point clouds of "Crawl" and "Long Jump". 
Furthermore, we present sports motions captured in large-scale multi-person sports scenes. These motions have greater challenges due to more self-occlusion, as defined by the LIP~\cite{ren2023lidar} benchmark. In FreeMotion, our sports motions exhibit more severe issues of external occlusion, resulting in point clouds with only a few points or even no points due to occlusion caused by other actors. This can be observed in the middle point clouds of "Football" and "Ping-Pong".
Lastly, we also capture data in various daily walkway scenes, including crossing obstacles, carrying shoulder bags and backpacks, etc. These types of motion frequently occur in daily life, and the most challenging issues are lots of human-irrelevant noise and occlusion caused by the interactive objects. 

%% file: sup_sections/more_visul.tex
\section{More experiments}
\label{sec:more_ex}
In this section, we show more experiments and more qualitative results of our ablation study.

\begin{table}[ht!]\scriptsize
\centering
\caption{Quantitative evaluation with fine-tune LiveHPS on each dataset. For the datasets with incomplete annotations, we use N/A for the metrics without annotation.}
\resizebox{\linewidth}{!}{
\begin{tabular}{cccccc}
\toprule
\multirow{2}{*}{} & \multicolumn{5}{c}{LiveHPS}                             \\
\cmidrule(r){2-6}
& J/V Err(P)$\downarrow$ & J/V Err(PS)$\downarrow$ & J/V Err(PST)$\downarrow$ & And Err$\downarrow$ & SUCD$\downarrow$ \\
\midrule
LiDARHuman26M~\cite{li2022lidarcap}& 65.34/81.17 & - & - & 18.69 & 2.62 \\
LIPD~\cite{ren2023lidar}& 56.71/70.53& - & - & 13.97 & 1.75 \\
CIMI4D~\cite{yan2023cimi4d}& 85.82/105.77& 86.33/106.14 & - & 25.02& 3.65\\
\midrule
FreeMotion& 66.38/80.10 & 66.68/80.52 & 116.72/125.46&15.59 &2.71\\
\bottomrule
\end{tabular}}
\label{tab:extension}
\vspace{-4ex}
\end{table}

our LiveHPS can provide high-quality results, notably in datasets like CIMI4D, where our results often align more closely with point clouds than global ground truth mesh. With SUCD metric, we can sift through a bunch of high-quality estimations as pseudo-labels to supplement lacking SMPL annotations in LiDARHuman26M~\cite{li2022lidarcap}, LIPD~\cite{ren2023lidar} and parts of our FreeMotion, particularly in outdoor multi-person scenes lacking dense IMU annotation.  These pseudo-labels are then incorporated into our training set to fine-tune our LiveHPS. Tab.\ref{tab:extension} displays the enhanced performance post fine-tuning, affirming LiveHPS's effectiveness and robustness, and its potential as an automated human pose and shape annotation tool.

\begin{figure}[ht!]
	\centering
	\includegraphics[width=\linewidth]{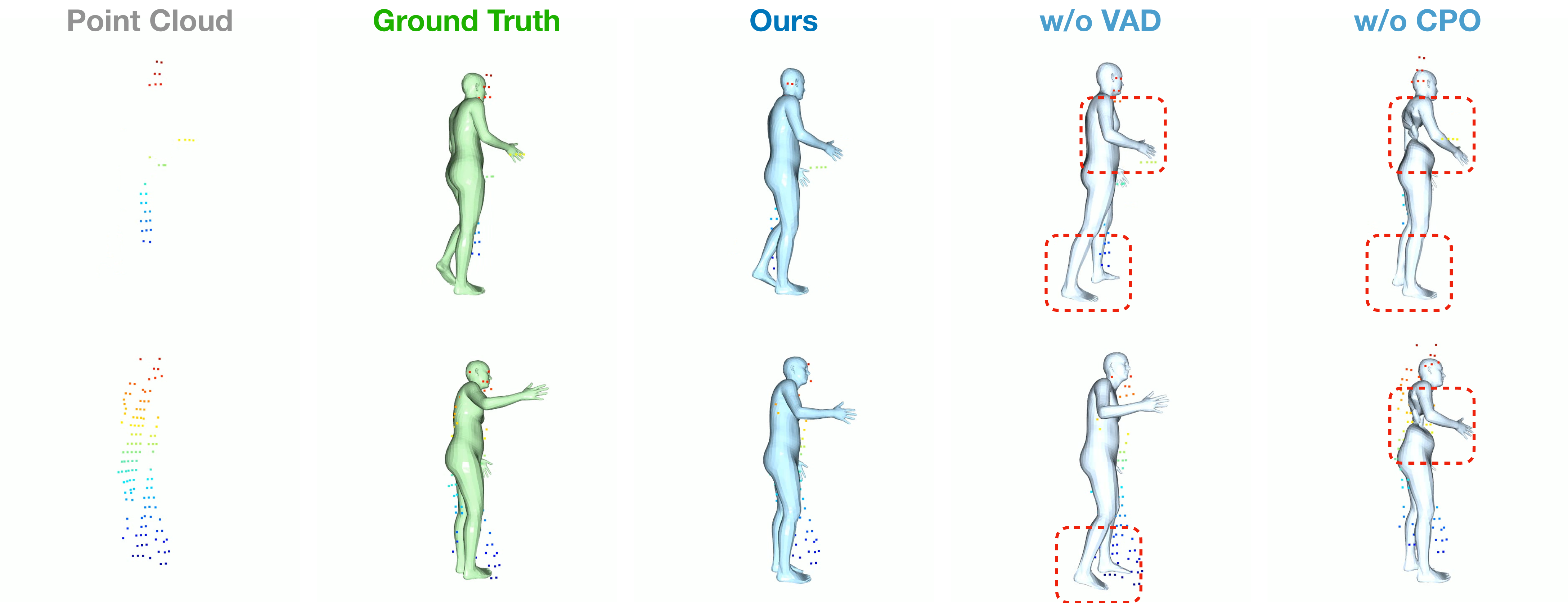}
	\caption{Qualitative evaluation on our network modules.}
	\label{fig:ab1}
	\vspace{-3ex}
\end{figure}

\begin{figure}[ht!]
	\centering
	\includegraphics[width=\linewidth]{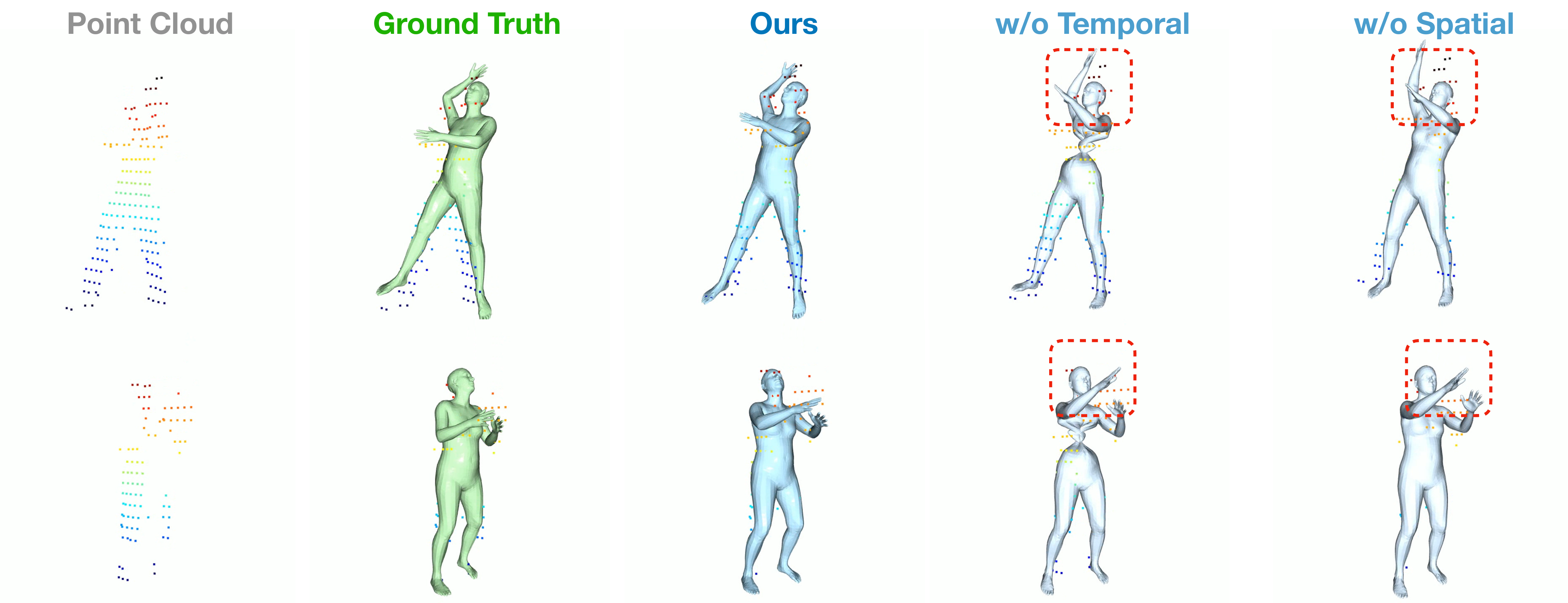}
	\caption{Qualitative evaluation on different optimization configurations of CPO module.}
	\label{fig:ab2}
	\vspace{-3ex}
\end{figure}

\begin{figure}[ht!]
	\centering
	\includegraphics[width=\linewidth]{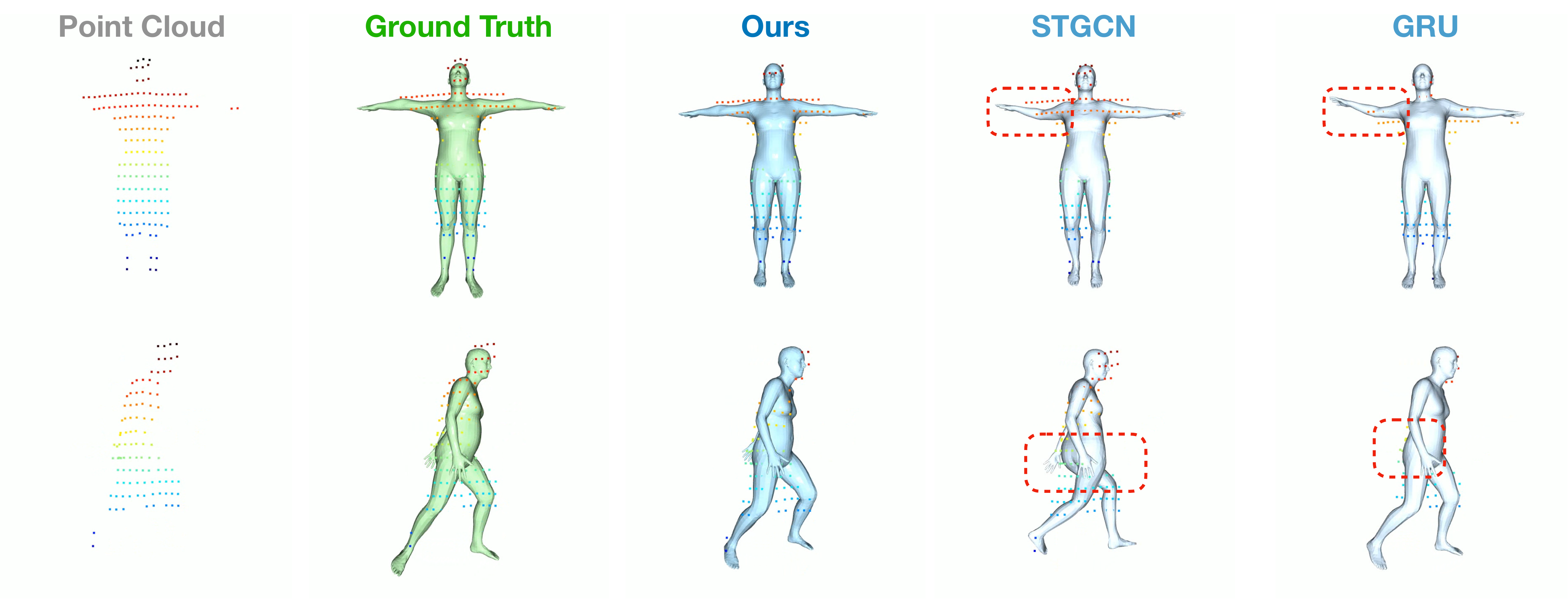}
	\caption{Qualitative evaluation on our attention-based Inverse Kinematic Solver.}
	\label{fig:ab3}
	\vspace{-3ex}
\end{figure}

\begin{figure}[ht!]
	\centering
	\includegraphics[width=\linewidth]{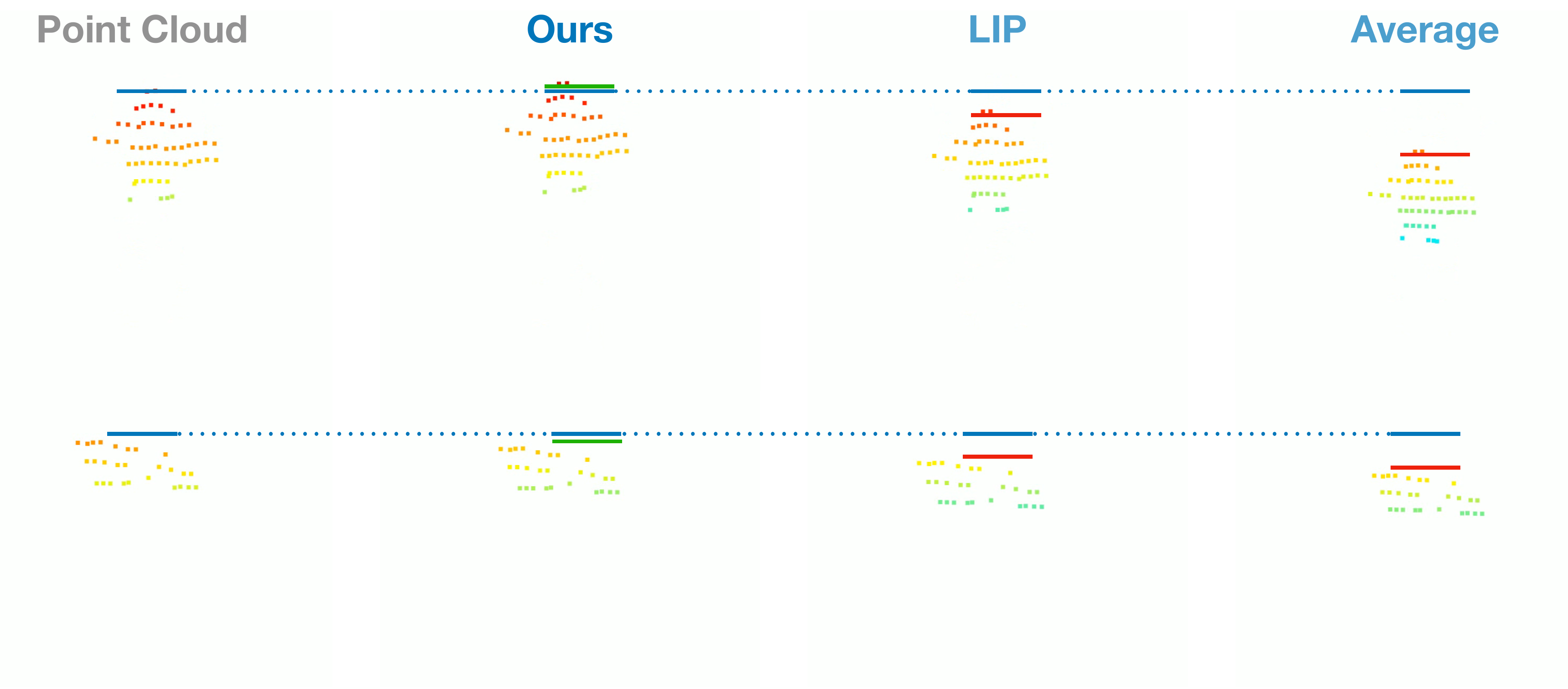}
	\caption{Qualitative evaluation on our network modules. The blue line means the height of point cloud with ground-truth translation, the green line means the height of point cloud with our predicted translation, and the red line means the point cloud with LIP predicted translation or average locations.}
	\label{fig:ab4}
	\vspace{-3ex}
\end{figure}

We show the qualitative evaluation of the ablation study, the red box means the incorrect local motion. As the Figure.~\ref{fig:ab1} shows, the network without consecutive pose optimizer module is unable to capture coherence features in temporal and spatial, which causes the big mistake in rotation estimation even the joint positions are basically correct. And the network without vertex-guided adaptive distillation module also performs bad in local motions when the input point cloud is occlusion, such as the foots and hands. As shown in Figure.~\ref{fig:ab2}, the network without temporal information also cause the same mistake in rotation estimation. Our method performs better especially in local motions, such as the head and hands. As shown in Figure.~\ref{fig:ab3}, when the point cloud is occlusion, our method can maintain stability, while others clearly affected by the occlusion of point clouds, such as the right hands in first row, the legs of STGCN result and left hand of GRU result in second row. Finally, we evaluate the translation estimation, the MOVIN provides the velocity estimation method for translation estimation, which causes severe cumulative errors, and other methods all can provide a basically correct result. In order to observe the accuracy of translation prediction more accurately, we subtract the predicted translation from the original point cloud to obtain the point cloud in the origin point. As the Figure.~\ref{fig:ab4} shows, our method still maintain stability when the actor is squatting.

%% file: sup_sections/more_results.tex
\section{More Results of LiveHPS}
\label{Sec:more_r}
In this section, we present additional multi-frame panoptic results of LiveHPS, which further illustrate the effectiveness and generalization capability of our method for estimating the accurate local pose in diverse challenging scenarios, encompassing indoor, outdoor, and night scenes.

\begin{figure*}[t]
	\centering
	\includegraphics[width=\linewidth]{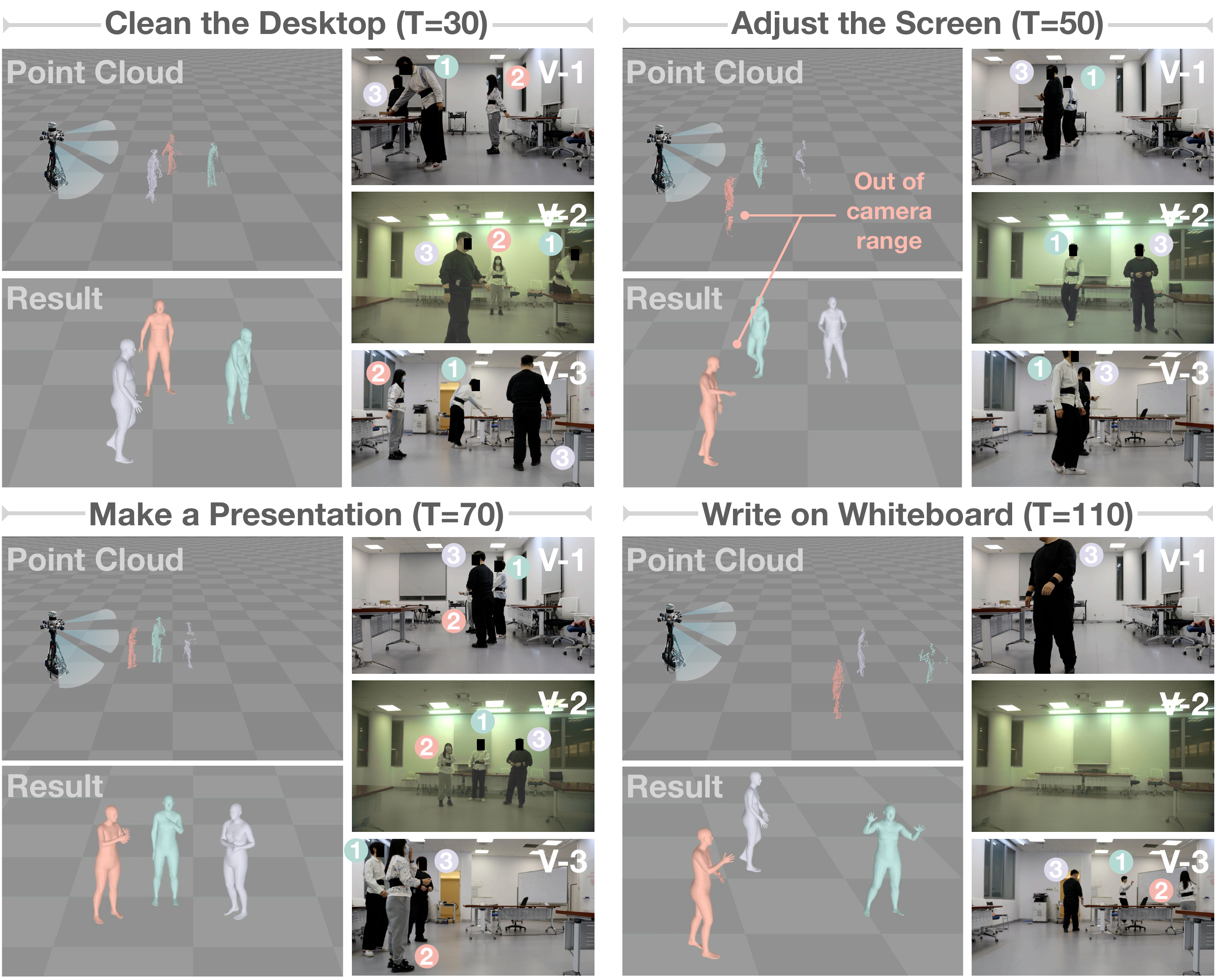}
	\caption{Sequential result of LiveHPS in indoor scene. We select four time points to show the indoor speech motion. The "T" means the timestamp, and the time unit is seconds. For each timestamp, we show point cloud, the corresponding result of LiveHPS and the image reference, "V-x" with same color means the three views of camera in the same timestamp. Digital labels with different colors on the image corresponding the point clouds with the same color. Note, for showing more details, we zoom in the results.}
	\label{fig:room}
	\vspace{-3ex}
\end{figure*}

\textbf{Indoor scene.} Most existing LiDAR-based mocap datasets~\cite{ren2023lidar,li2022lidarcap} primarily focus on capturing data in outdoor scenes to leverage the wide coverage capabilities of LiDAR sensors. Moreover, indoor data often present challenges such as lots of noise and occlusion in the point cloud caused by surrounding complex furniture in limited ranges. As shown in Figure.~\ref{fig:room}, three actors are engaged in indoor speech motions, involving tasks such as cleaning the desktop, adjusting the screen, giving a presentation, and writing on a whiteboard. We also display image reference in three views, in some case, actors may fall outside the camera's range. External occlusions arise due to the presence of other actors (purple actor in "Adjust the Screen") and indoor facilities (cyan actor in "Write on Whiteboard"). Despite these challenges, our method demonstrates reliable performance in handling occluded point clouds, showing its robustness in complex indoor environments.

\begin{figure*}[t]
	\centering
	\includegraphics[width=\linewidth]{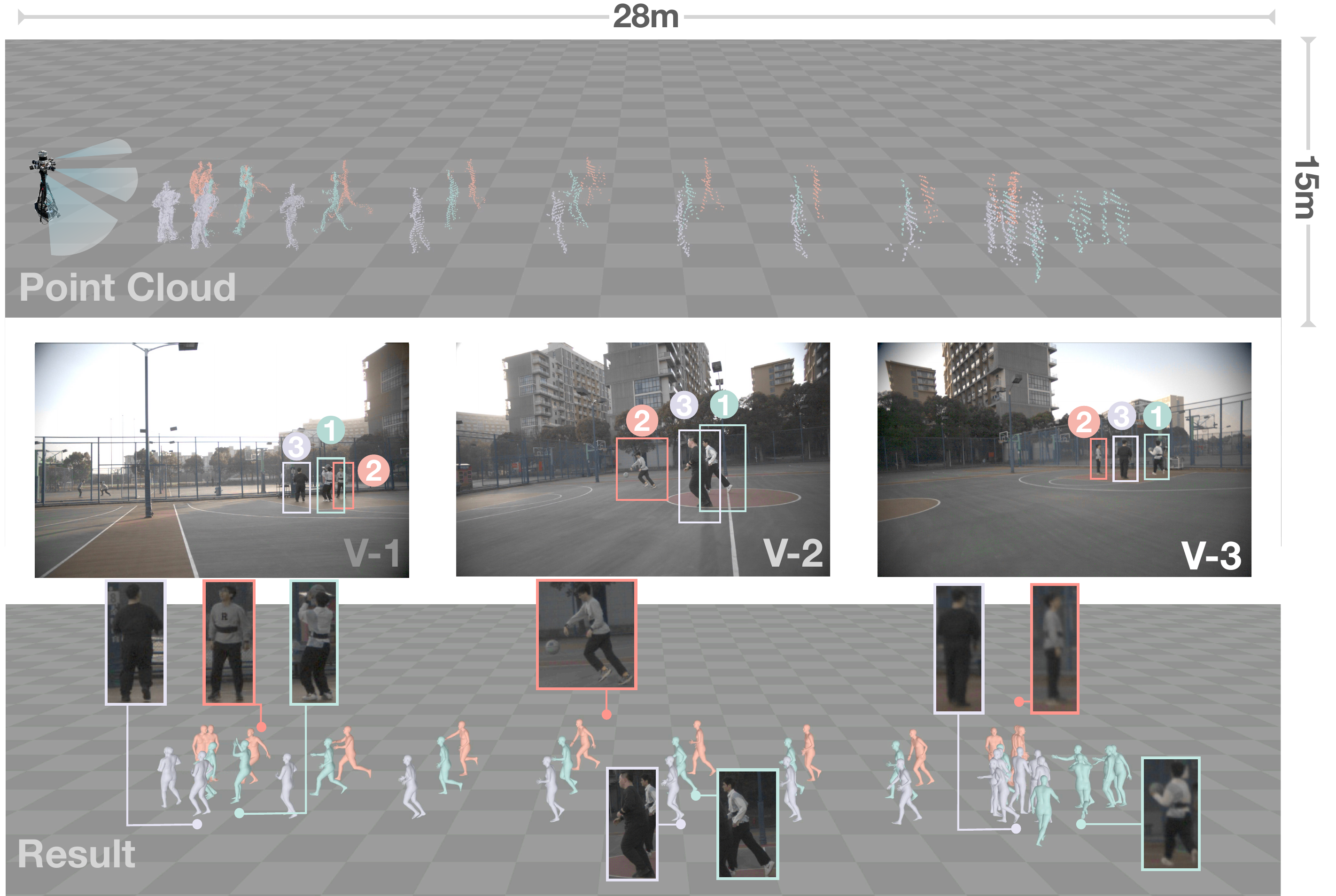}
	\caption{Sequential result of LiveHPS in outdoor scene. We show the sequential point cloud and results with time step of 10 seconds. For showing the image reference in three views, we select three different timestamp for each view, the "V-x" with different colors means different timestamp in view-x(x=1,2,3). Digital labels with different colors on the image corresponding the point clouds with the same color.}
	\label{fig:bk}
	\vspace{-3ex}
\end{figure*}

\textbf{Outdoor scene.} Benefiting from long-range sensing properties of LiDAR, FreeMotion can capture multi-person motions in various large-scale scenes. As shown in Figure.~\ref{fig:bk}, we collect the data in the basketball court measuring 28 meters in length and 15 meters in width. The sequence shows a fast-break tactic involving three actors, and external occlusion primarily occurs during instances when the actors scramble for the basketball, as observed in the right part of the sequence. Despite the challenging conditions, our method exhibits excellent performance even in situations where the point cloud becomes extremely sparse due to actors being distant from the LiDAR sensor or being occluded by others. These results highlight the potential applicability of LiveHPS in sports events.

\begin{figure*}[t]
	\centering
	\includegraphics[width=\linewidth]{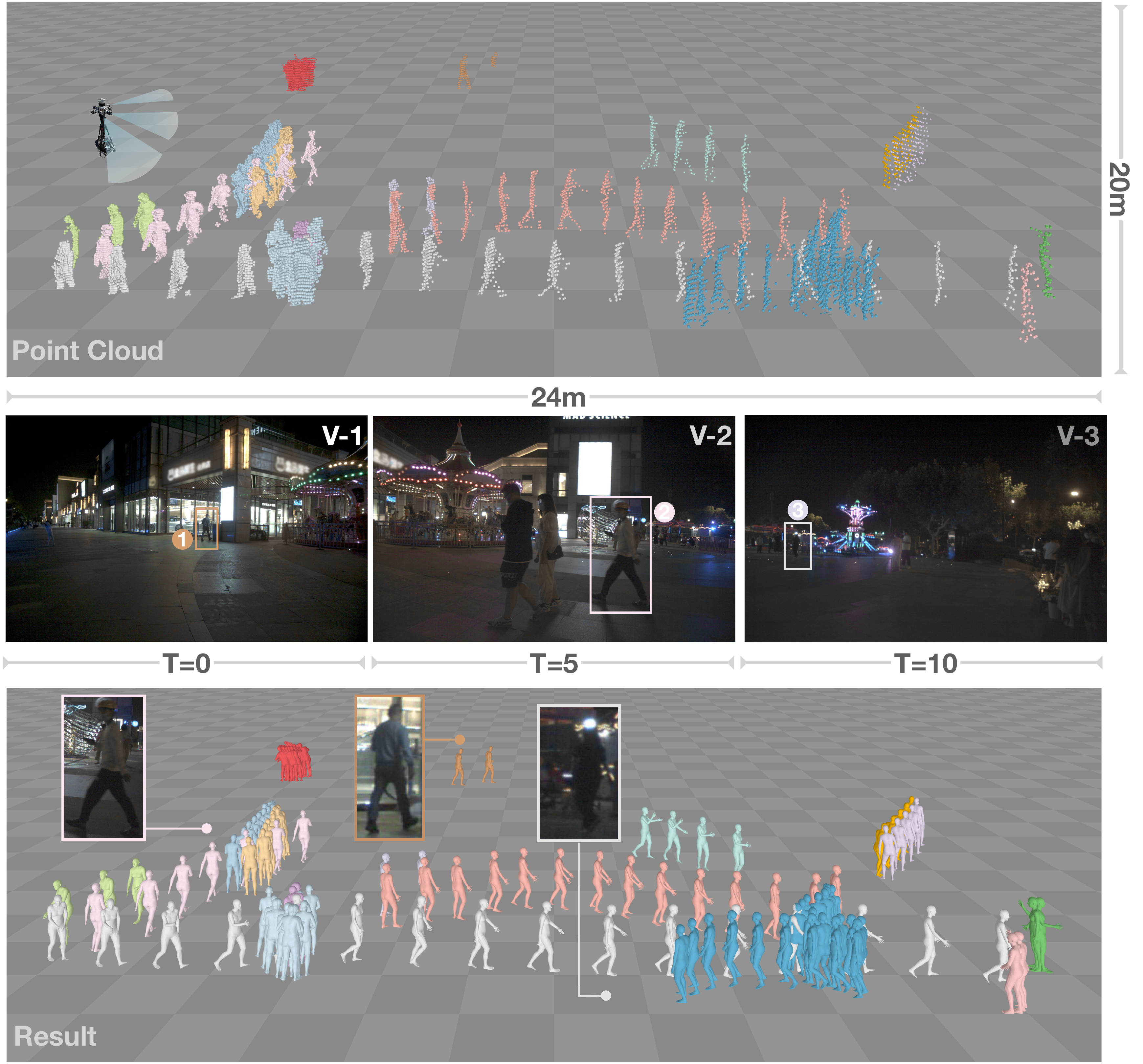}
	\caption{Sequential result of LiveHPS in night scene. The time step of the sequential point cloud and results is 10 seconds. The "V-x" with different colors means different timestamp in view-x(x=1,2,3). T means the timestamp of the image and the time unit is seconds. Digital labels with different colors on the image corresponding the point clouds with the same color.}
	\label{fig:ct}
	\vspace{-3ex}
\end{figure*}

\textbf{Night scene.} LiDAR sensors operate independently of environmental lighting conditions, which can work in the completely dark environments. In Figure.~\ref{fig:ct}, we present the sequential results of LiveHPS captured on the square at night. The scene depicts a densely populated area with complex terrain. The LiDAR can capture seventeen persons in the region measuring 24 meters in length and 20 meters in width. In contrast, the image reference is significantly unclear due to the absence of adequate lighting. Nevertheless, our method demonstrates reliable performance in this challenging scene, demonstrating that LiveHPS can be applied for arbitrary real-world scenes.

%% file: main.bbl
\begin{thebibliography}{70}
\providecommand{\natexlab}[1]{#1}
\providecommand{\url}[1]{\texttt{#1}}
\expandafter\ifx\csname urlstyle\endcsname\relax
  \providecommand{\doi}[1]{doi: #1}\else
  \providecommand{\doi}{doi: \begingroup \urlstyle{rm}\Url}\fi

\bibitem[eas(2021)]{easymocap}
Easymocap - make human motion capture easier.
\newblock Github, 2021.

\bibitem[Amin et~al.(2009)Amin, Andriluka, Rohrbach, and Schiele]{AminARS2009}
Sikander Amin, Mykhaylo Andriluka, Marcus Rohrbach, and Bernt Schiele.
\newblock Multi-view pictorial structures for {3D} human pose estimation.
\newblock In \emph{BMVC}, 2009.

\bibitem[Baak et~al.(2011)Baak, M\"{u}ller, Bharaj, Seidel, and Theobalt]{Baak:2011}
Andreas Baak, Meinard M\"{u}ller, Gaurav Bharaj, Hans-Peter Seidel, and Christian Theobalt.
\newblock A data-driven approach for real-time full body pose reconstruction from a depth camera.
\newblock In \emph{ICCV}, 2011.

\bibitem[Behley et~al.(2019)Behley, Garbade, Milioto, Quenzel, Behnke, Stachniss, and Gall]{behley2019semantickitti}
Jens Behley, Martin Garbade, Andres Milioto, Jan Quenzel, Sven Behnke, Cyrill Stachniss, and Jurgen Gall.
\newblock Semantickitti: A dataset for semantic scene understanding of lidar sequences.
\newblock In \emph{Proceedings of the IEEE/CVF international conference on computer vision}, pages 9297--9307, 2019.

\bibitem[Besl and McKay(1992)]{besl1992method}
Paul~J Besl and Neil~D McKay.
\newblock Method for registration of 3-d shapes.
\newblock In \emph{Sensor fusion IV: control paradigms and data structures}, pages 586--606. Spie, 1992.

\bibitem[Bogo et~al.(2016)Bogo, Kanazawa, Lassner, Gehler, Romero, and Black]{bogo2016keep}
Federica Bogo, Angjoo Kanazawa, Christoph Lassner, Peter Gehler, Javier Romero, and Michael~J Black.
\newblock Keep it smpl: Automatic estimation of 3d human pose and shape from a single image.
\newblock In \emph{Computer Vision--ECCV 2016: 14th European Conference, Amsterdam, The Netherlands, October 11-14, 2016, Proceedings, Part V 14}, pages 561--578. Springer, 2016.

\bibitem[Burenius et~al.(2013)Burenius, Sullivan, and Carlsson]{BurenSC2013}
Magnus Burenius, Josephine Sullivan, and Stefan Carlsson.
\newblock {3D} pictorial structures for multiple view articulated pose estimation.
\newblock In \emph{CVPR}, 2013.

\bibitem[Cai et~al.(2022)Cai, Ren, Zeng, Lin, Yu, Wang, Fan, Gao, Yu, Pan, et~al.]{cai2022humman}
Zhongang Cai, Daxuan Ren, Ailing Zeng, Zhengyu Lin, Tao Yu, Wenjia Wang, Xiangyu Fan, Yang Gao, Yifan Yu, Liang Pan, et~al.
\newblock Humman: Multi-modal 4d human dataset for versatile sensing and modeling.
\newblock In \emph{European Conference on Computer Vision}, pages 557--577. Springer, 2022.

\bibitem[Cai et~al.(2023)Cai, Pan, Wei, Yin, Hong, Zhang, Loy, Yang, and Liu]{cai2023pointhps}
Zhongang Cai, Liang Pan, Chen Wei, Wanqi Yin, Fangzhou Hong, Mingyuan Zhang, Chen~Change Loy, Lei Yang, and Ziwei Liu.
\newblock Pointhps: Cascaded 3d human pose and shape estimation from point clouds.
\newblock \emph{arXiv preprint arXiv:2308.14492}, 2023.

\bibitem[Cong et~al.(2022{\natexlab{a}})Cong, Zhu, Qiao, Ren, Peng, Hou, Xu, Yang, Manocha, and Ma]{Cong_2022_CVPR}
Peishan Cong, Xinge Zhu, Feng Qiao, Yiming Ren, Xidong Peng, Yuenan Hou, Lan Xu, Ruigang Yang, Dinesh Manocha, and Yuexin Ma.
\newblock Stcrowd: A multimodal dataset for pedestrian perception in crowded scenes.
\newblock In \emph{CVPR}, pages 19608--19617, 2022{\natexlab{a}}.

\bibitem[Cong et~al.(2022{\natexlab{b}})Cong, Zhu, Qiao, Ren, Peng, Hou, Xu, Yang, Manocha, and Ma]{cong2022stcrowd}
Peishan Cong, Xinge Zhu, Feng Qiao, Yiming Ren, Xidong Peng, Yuenan Hou, Lan Xu, Ruigang Yang, Dinesh Manocha, and Yuexin Ma.
\newblock Stcrowd: A multimodal dataset for pedestrian perception in crowded scenes.
\newblock \emph{arXiv preprint arXiv:2204.01026}, 2022{\natexlab{b}}.

\bibitem[Dai et~al.(2022)Dai, Lin, Wen, Shen, Xu, Yu, Ma, and Wang]{dai2022hsc4d}
Yudi Dai, Yitai Lin, Chenglu Wen, Siqi Shen, Lan Xu, Jingyi Yu, Yuexin Ma, and Cheng Wang.
\newblock Hsc4d: Human-centered 4d scene capture in large-scale indoor-outdoor space using wearable imus and lidar.
\newblock In \emph{CVPR}, pages 6792--6802, 2022.

\bibitem[Dai et~al.(2023)Dai, Lin, Lin, Wen, Xu, Yi, Shen, Ma, and Wang]{dai2023sloper4d}
Yudi Dai, Yitai Lin, Xiping Lin, Chenglu Wen, Lan Xu, Hongwei Yi, Siqi Shen, Yuexin Ma, and Cheng Wang.
\newblock Sloper4d: A scene-aware dataset for global 4d human pose estimation in urban environments.
\newblock \emph{arXiv preprint arXiv:2303.09095}, 2023.

\bibitem[Elhayek et~al.(2015)Elhayek, de~Aguiar, Jain, Tompson, Pishchulin, Andriluka, Bregler, Schiele, and Theobalt]{ElhayAJTPABST2015}
Ahmed Elhayek, Edilson de Aguiar, Arjun Jain, Jonathan Tompson, Leonid Pishchulin, Mykhaylo Andriluka, Chris Bregler, Bernt Schiele, and Christian Theobalt.
\newblock Efficient {ConvNet}-based marker-less motion capture in general scenes with a low number of cameras.
\newblock In \emph{CVPR}, 2015.

\bibitem[Ester et~al.(1996)Ester, Kriegel, Sander, Xu, et~al.]{ester1996density}
Martin Ester, Hans-Peter Kriegel, J{\"o}rg Sander, Xiaowei Xu, et~al.
\newblock A density-based algorithm for discovering clusters in large spatial databases with noise.
\newblock In \emph{kdd}, pages 226--231, 1996.

\bibitem[Guo et~al.(2018)Guo, Taylor, Fanello, Tagliasacchi, Dou, Davidson, Kowdle, and Izadi]{guoTwinFusion}
Kaiwen Guo, Jonathan Taylor, Sean Fanello, Andrea Tagliasacchi, Mingsong Dou, Philip Davidson, Adarsh Kowdle, and Shahram Izadi.
\newblock Twinfusion: High framerate non-rigid fusion through fast correspondence tracking.
\newblock In \emph{3DV}, pages 596--605, 2018.

\bibitem[Habermann et~al.(2019)Habermann, Xu, Zollh\"{o}fer, Pons-Moll, and Theobalt]{LiveCap2019tog}
Marc Habermann, Weipeng Xu, Michael Zollh\"{o}fer, Gerard Pons-Moll, and Christian Theobalt.
\newblock Livecap: Real-time human performance capture from monocular video.
\newblock \emph{ACM Transactions on Graphics (TOG)}, 38\penalty0 (2):\penalty0 14:1--14:17, 2019.

\bibitem[Habermann et~al.(2020)Habermann, Xu, Zollhofer, Pons-Moll, and Theobalt]{DeepCap_CVPR2020}
Marc Habermann, Weipeng Xu, Michael Zollhofer, Gerard Pons-Moll, and Christian Theobalt.
\newblock Deepcap: Monocular human performance capture using weak supervision.
\newblock In \emph{CVPR}, 2020.

\bibitem[He et~al.(2021)He, Pang, Chen, Liang, Wu, Ma, and Xu]{challencap}
Yannan He, Anqi Pang, Xin Chen, Han Liang, Minye Wu, Yuexin Ma, and Lan Xu.
\newblock Challencap: Monocular 3d capture of challenging human performances using multi-modal references.
\newblock In \emph{CVPR}, pages 11400--11411, 2021.

\bibitem[{Huang} et~al.(2017){Huang}, {Bogo}, {Lassner}, {Kanazawa}, {Gehler}, {Romero}, {Akhter}, and {Black}]{TAM_3DV2017}
Y. {Huang}, F. {Bogo}, C. {Lassner}, A. {Kanazawa}, P.~V. {Gehler}, J. {Romero}, I. {Akhter}, and M.~J. {Black}.
\newblock Towards accurate marker-less human shape and pose estimation over time.
\newblock In \emph{3DV}, pages 421--430, 2017.

\bibitem[Huang et~al.(2018)Huang, Kaufmann, Aksan, Black, Hilliges, and Pons-Moll]{huang2018DIP}
Yinghao Huang, Manuel Kaufmann, Emre Aksan, Michael~J Black, Otmar Hilliges, and Gerard Pons-Moll.
\newblock Deep inertial poser: Learning to reconstruct human pose from sparse inertial measurements in real time.
\newblock \emph{ACM Transactions on Graphics (TOG)}, 37\penalty0 (6):\penalty0 1--15, 2018.

\bibitem[Ionescu et~al.(2013)Ionescu, Papava, Olaru, and Sminchisescu]{ionescu2013human3}
Catalin Ionescu, Dragos Papava, Vlad Olaru, and Cristian Sminchisescu.
\newblock Human3. 6m: Large scale datasets and predictive methods for 3d human sensing in natural environments.
\newblock \emph{TPAMI}, 36\penalty0 (7):\penalty0 1325--1339, 2013.

\bibitem[Jang et~al.(2023)Jang, Yang, Jang, Choi, Jin, and Lee]{jang2023movin}
Deok-Kyeong Jang, Dongseok Yang, Deok-Yun Jang, Byeoli Choi, Taeil Jin, and Sung-Hee Lee.
\newblock Movin: Real-time motion capture using a single lidar.
\newblock \emph{arXiv preprint arXiv:2309.09314}, 2023.

\bibitem[Joo et~al.(2015)Joo, Liu, Tan, Gui, Nabbe, Matthews, Kanade, Nobuhara, and Sheikh]{JooLTGNMKNS2015}
Hanbyul Joo, Hao Liu, Lei Tan, Lin Gui, Bart Nabbe, Iain Matthews, Takeo Kanade, Shohei Nobuhara, and Yaser Sheikh.
\newblock Panoptic studio: A massively multiview system for social motion capture.
\newblock In \emph{ICCV}, 2015.

\bibitem[Kanazawa et~al.(2018)Kanazawa, Black, Jacobs, and Malik]{HMR18}
Angjoo Kanazawa, Michael~J. Black, David~W. Jacobs, and Jitendra Malik.
\newblock End-to-end recovery of human shape and pose.
\newblock In \emph{CVPR}, 2018.

\bibitem[Kanazawa et~al.(2019)Kanazawa, Zhang, Felsen, and Malik]{Kanazawa_2019CVPR}
Angjoo Kanazawa, Jason~Y. Zhang, Panna Felsen, and Jitendra Malik.
\newblock Learning 3d human dynamics from video.
\newblock In \emph{CVPR}, 2019.

\bibitem[Kim et~al.(2019)Kim, Ramanagopal, Barto, Yu, Rosaen, Goumas, Vasudevan, and Johnson-Roberson]{kim2019pedx}
Wonhui Kim, Manikandasriram~Srinivasan Ramanagopal, Charles Barto, Ming-Yuan Yu, Karl Rosaen, Nick Goumas, Ram Vasudevan, and Matthew Johnson-Roberson.
\newblock Pedx: Benchmark dataset for metric 3-d pose estimation of pedestrians in complex urban intersections.
\newblock \emph{IRAL}, 4\penalty0 (2):\penalty0 1940--1947, 2019.

\bibitem[Kocabas et~al.(2020)Kocabas, Athanasiou, and Black]{VIBE_CVPR2020}
Muhammed Kocabas, Nikos Athanasiou, and Michael~J. Black.
\newblock Vibe: Video inference for human body pose and shape estimation.
\newblock In \emph{CVPR}, 2020.

\bibitem[Kocabas et~al.(2021)Kocabas, Huang, Hilliges, and Black]{PARE_ICCV2021}
Muhammed Kocabas, Chun-Hao~P. Huang, Otmar Hilliges, and Michael~J. Black.
\newblock Pare: Part attention regressor for 3d human body estimation.
\newblock In \emph{ICCV}, pages 11127--11137, 2021.

\bibitem[Kolotouros et~al.(2019)Kolotouros, Pavlakos, and Daniilidis]{Kolotouros_2019_CVPR}
Nikos Kolotouros, Georgios Pavlakos, and Kostas Daniilidis.
\newblock Convolutional mesh regression for single-image human shape reconstruction.
\newblock In \emph{CVPR}, 2019.

\bibitem[Kolotouros et~al.(2021)Kolotouros, Pavlakos, Jayaraman, and Daniilidis]{PHMR_ICCV2021}
Nikos Kolotouros, Georgios Pavlakos, Dinesh Jayaraman, and Kostas Daniilidis.
\newblock Probabilistic modeling for human mesh recovery.
\newblock In \emph{ICCV}, pages 11605--11614, 2021.

\bibitem[Lassner et~al.(2017)Lassner, Romero, Kiefel, Bogo, Black, and Gehler]{Lassner17}
Christoph Lassner, Javier Romero, Martin Kiefel, Federica Bogo, Michael~J Black, and Peter~V Gehler.
\newblock Unite the people: Closing the loop between 3d and 2d human representations.
\newblock In \emph{CVPR}, pages 6050--6059, 2017.

\bibitem[Li et~al.(2022)Li, Zhang, Wang, Shen, Wen, Ma, Xu, Yu, and Wang]{li2022lidarcap}
Jialian Li, Jingyi Zhang, Zhiyong Wang, Siqi Shen, Chenglu Wen, Yuexin Ma, Lan Xu, Jingyi Yu, and Cheng Wang.
\newblock Lidarcap: Long-range marker-less 3d human motion capture with lidar point clouds.
\newblock \emph{arXiv preprint arXiv:2203.14698}, 2022.

\bibitem[Li et~al.(2021)Li, Yang, Ross, and Kanazawa]{li2021learn}
Ruilong Li, Shan Yang, David~A. Ross, and Angjoo Kanazawa.
\newblock Ai choreographer: Music conditioned 3d dance generation with aist++, 2021.

\bibitem[Loper et~al.(2015)Loper, Mahmood, Romero, Pons-Moll, and Black]{SMPL2015}
Matthew Loper, Naureen Mahmood, Javier Romero, Gerard Pons-Moll, and Michael~J. Black.
\newblock Smpl: A skinned multi-person linear model.
\newblock \emph{ACM Trans. Graph.}, 34\penalty0 (6):\penalty0 248:1--248:16, 2015.

\bibitem[Luo et~al.(2021)Luo, Hachiuma, Yuan, and Kitani]{luo2021dynamics}
Zhengyi Luo, Ryo Hachiuma, Ye Yuan, and Kris Kitani.
\newblock Dynamics-regulated kinematic policy for egocentric pose estimation.
\newblock \emph{Advances in Neural Information Processing Systems}, 34, 2021.

\bibitem[Mahmood et~al.(2019)Mahmood, Ghorbani, Troje, Pons-Moll, and Black]{AMASS_ICCV2019}
Naureen Mahmood, Nima Ghorbani, Nikolaus~F. Troje, Gerard Pons-Moll, and Michael~J. Black.
\newblock Amass: Archive of motion capture as surface shapes.
\newblock In \emph{ICCV}, 2019.

\bibitem[Mehta et~al.(2017)Mehta, Rhodin, Casas, Fua, Sotnychenko, Xu, and Theobalt]{mehta2017monocular}
Dushyant Mehta, Helge Rhodin, Dan Casas, Pascal Fua, Oleksandr Sotnychenko, Weipeng Xu, and Christian Theobalt.
\newblock Monocular 3d human pose estimation in the wild using improved cnn supervision.
\newblock In \emph{3DV}, pages 506--516. IEEE, 2017.

\bibitem[Noitom()]{noitom}
Noitom.
\newblock {Noitom Motion Capture Systems}.
\newblock \url{https://www.noitom.com/}, 2015.

\bibitem[OptiTrack()]{optitrack}
OptiTrack.
\newblock {OptiTrack Motion Capture Systems}.
\newblock \url{https://www.optitrack.com/}, 2009.

\bibitem[Pavlakos et~al.(2017)Pavlakos, Zhou, Derpanis, and Daniilidis]{Pavlakos17}
Georgios Pavlakos, Xiaowei Zhou, Konstantinos~G Derpanis, and Kostas Daniilidis.
\newblock Harvesting multiple views for marker-less 3d human pose annotations.
\newblock In \emph{CVPR}, 2017.

\bibitem[Peng et~al.(2023)Peng, Zhu, and Ma]{peng2022cl3d}
Xidong Peng, Xinge Zhu, and Yuexin Ma.
\newblock Cl3d: Unsupervised domain adaptation for cross-lidar 3d detection.
\newblock \emph{AAAI}, 2023.

\bibitem[RealityCapture()]{rc}
RealityCapture.
\newblock {Capturing Reality}.
\newblock \url{https://www.capturingreality.com/}, 2023.

\bibitem[Rempe et~al.(2021)Rempe, Birdal, Hertzmann, Yang, Sridhar, and Guibas]{HUMOR_ICCV2021}
Davis Rempe, Tolga Birdal, Aaron Hertzmann, Jimei Yang, Srinath Sridhar, and Leonidas~J. Guibas.
\newblock Humor: 3d human motion model for robust pose estimation.
\newblock In \emph{ICCV}, pages 11488--11499, 2021.

\bibitem[Ren et~al.(2023)Ren, Zhao, He, Cong, Liang, Yu, Xu, and Ma]{ren2023lidar}
Yiming Ren, Chengfeng Zhao, Yannan He, Peishan Cong, Han Liang, Jingyi Yu, Lan Xu, and Yuexin Ma.
\newblock Lidar-aid inertial poser: Large-scale human motion capture by sparse inertial and lidar sensors.
\newblock \emph{TVCG}, 2023.

\bibitem[Rhodin et~al.(2015)Rhodin, Robertini, Richardt, Seidel, and Theobalt]{RhodiRRST2015}
Helge Rhodin, Nadia Robertini, Christian Richardt, Hans-Peter Seidel, and Christian Theobalt.
\newblock A versatile scene model with differentiable visibility applied to generative pose estimation.
\newblock In \emph{ICCV}, 2015.

\bibitem[Robertini et~al.(2016)Robertini, Casas, Rhodin, Seidel, and Theobalt]{Robertini:2016}
Nadia Robertini, Dan Casas, Helge Rhodin, Hans-Peter Seidel, and Christian Theobalt.
\newblock Model-based outdoor performance capture.
\newblock In \emph{3DV}, 2016.

\bibitem[Shotton et~al.(2011)Shotton, Fitzgibbon, Cook, Sharp, Finocchio, Moore, Kipman, and Blake]{Shotton:2011}
Jamie Shotton, Andrew Fitzgibbon, Mat Cook, Toby Sharp, Mark Finocchio, Richard Moore, Alex Kipman, and Andrew Blake.
\newblock Real-time human pose recognition in parts from single depth images.
\newblock In \emph{CVPR}, 2011.

\bibitem[Sigal et~al.(2010)Sigal, B{\u{a}}lan, and Black]{SigalBB2010}
Leonid Sigal, Alexandru~O. B{\u{a}}lan, and Michael~J. Black.
\newblock {HumanEva}: Synchronized video and motion capture dataset and baseline algorithm for evaluation of articulated human motion.
\newblock \emph{IJCV}, 2010.

\bibitem[Simon et~al.(2017)Simon, Joo, Matthews, and Sheikh]{Simon17}
Tomas Simon, Hanbyul Joo, Iain Matthews, and Yaser Sheikh.
\newblock Hand keypoint detection in single images using multiview bootstrapping.
\newblock In \emph{CVPR}, 2017.

\bibitem[Varol et~al.(2017)Varol, Romero, Martin, Mahmood, Black, Laptev, and Schmid]{varol2017learning}
Gul Varol, Javier Romero, Xavier Martin, Naureen Mahmood, Michael~J Black, Ivan Laptev, and Cordelia Schmid.
\newblock Learning from synthetic humans.
\newblock In \emph{Proceedings of the IEEE conference on computer vision and pattern recognition}, pages 109--117, 2017.

\bibitem[Vicon()]{VICON}
Vicon.
\newblock {Vicon Motion Capture Systems}.
\newblock \url{https://www.vicon.com/}, 2010.

\bibitem[Vlasic et~al.(2007)Vlasic, Adelsberger, Vannucci, Barnwell, Gross, Matusik, and Popovi{\'c}]{Vlasic2007}
Daniel Vlasic, Rolf Adelsberger, Giovanni Vannucci, John Barnwell, Markus Gross, Wojciech Matusik, and Jovan Popovi{\'c}.
\newblock Practical motion capture in everyday surroundings.
\newblock \emph{TOG}, 26\penalty0 (3):\penalty0 35--es, 2007.

\bibitem[Von~Marcard et~al.(2017)Von~Marcard, Rosenhahn, Black, and Pons-Moll]{von2017SIP}
Timo Von~Marcard, Bodo Rosenhahn, Michael~J Black, and Gerard Pons-Moll.
\newblock Sparse inertial poser: Automatic 3d human pose estimation from sparse imus.
\newblock In \emph{Computer Graphics Forum}, pages 349--360. Wiley Online Library, 2017.

\bibitem[Von~Marcard et~al.(2018)Von~Marcard, Henschel, Black, Rosenhahn, and Pons-Moll]{von2018recovering}
Timo Von~Marcard, Roberto Henschel, Michael~J Black, Bodo Rosenhahn, and Gerard Pons-Moll.
\newblock Recovering accurate 3d human pose in the wild using imus and a moving camera.
\newblock In \emph{ECCV}, pages 601--617, 2018.

\bibitem[Vu et~al.(2022)Vu, Kim, Luu, Nguyen, and Yoo]{vu2022softgroup}
Thang Vu, Kookhoi Kim, Tung~M Luu, Thanh Nguyen, and Chang~D Yoo.
\newblock Softgroup for 3d instance segmentation on point clouds.
\newblock In \emph{Proceedings of the IEEE/CVF Conference on Computer Vision and Pattern Recognition}, pages 2708--2717, 2022.

\bibitem[Wei et~al.(2012)Wei, Zhang, and Chai]{Wei:2012}
Xiaolin Wei, Peizhao Zhang, and Jinxiang Chai.
\newblock Accurate realtime full-body motion capture using a single depth camera.
\newblock \emph{SIGGRAPH Asia}, 31\penalty0 (6):\penalty0 188:1--12, 2012.

\bibitem[XSENS()]{XSENS}
XSENS.
\newblock {Xsens Technologies B.V.}
\newblock \url{https://www.xsens.com/}, 2011.

\bibitem[Xu et~al.(2020)Xu, Xu, Golyanik, Habermann, Fang, and Theobalt]{EventCap_CVPR2020}
Lan Xu, Weipeng Xu, Vladislav Golyanik, Marc Habermann, Lu Fang, and Christian Theobalt.
\newblock Eventcap: Monocular 3d capture of high-speed human motions using an event camera.
\newblock In \emph{CVPR}, 2020.

\bibitem[Xu et~al.(2018)Xu, Chatterjee, Zollh\"{o}fer, Rhodin, Mehta, Seidel, and Theobalt]{MonoPerfCap}
Weipeng Xu, Avishek Chatterjee, Michael Zollh\"{o}fer, Helge Rhodin, Dushyant Mehta, Hans-Peter Seidel, and Christian Theobalt.
\newblock Monoperfcap: Human performance capture from monocular video.
\newblock \emph{ACM Transactions on Graphics (TOG)}, 37\penalty0 (2):\penalty0 27:1--27:15, 2018.

\bibitem[Xu et~al.(2023)Xu, Cong, Yao, Chen, Hou, Zhu, He, Yu, and Ma]{xu2023human}
Yiteng Xu, Peishan Cong, Yichen Yao, Runnan Chen, Yuenan Hou, Xinge Zhu, Xuming He, Jingyi Yu, and Yuexin Ma.
\newblock Human-centric scene understanding for 3d large-scale scenarios.
\newblock In \emph{Proceedings of the IEEE/CVF International Conference on Computer Vision}, pages 20349--20359, 2023.

\bibitem[Yan et~al.(2023)Yan, Wang, Dai, Shen, Wen, Xu, Ma, and Wang]{yan2023cimi4d}
Ming Yan, Xin Wang, Yudi Dai, Siqi Shen, Chenglu Wen, Lan Xu, Yuexin Ma, and Cheng Wang.
\newblock Cimi4d: A large multimodal climbing motion dataset under human-scene interactions.
\newblock \emph{arXiv preprint arXiv:2303.17948}, 2023.

\bibitem[Yi et~al.(2021)Yi, Zhou, and Xu]{yi2021transpose}
Xinyu Yi, Yuxiao Zhou, and Feng Xu.
\newblock Transpose: Real-time 3d human translation and pose estimation with six inertial sensors.
\newblock \emph{ACM Transactions on Graphics (TOG)}, 40\penalty0 (4):\penalty0 1--13, 2021.

\bibitem[Yi et~al.(2022)Yi, Zhou, Habermann, Shimada, Golyanik, Theobalt, and Xu]{PIPCVPR2022}
Xinyu Yi, Yuxiao Zhou, Marc Habermann, Soshi Shimada, Vladislav Golyanik, Christian Theobalt, and Feng Xu.
\newblock Physical inertial poser (pip): Physics-aware real-time human motion tracking from sparse inertial sensors.
\newblock In \emph{CVPR}, 2022.

\bibitem[Yin et~al.(2021)Yin, Zhou, and Kr{\"a}henb{\"u}hl]{Yin2020Centerbased3O}
Tianwei Yin, Xingyi Zhou, and Philipp Kr{\"a}henb{\"u}hl.
\newblock Center-based 3d object detection and tracking.
\newblock \emph{CVPR}, 2021.

\bibitem[Yu et~al.(2019)Yu, Zheng, Guo, Zhao, Dai, Li, Pons-Moll, and Liu]{DoubleFusion}
Tao Yu, Zerong Zheng, Kaiwen Guo, Jianhui Zhao, Qionghai Dai, Hao Li, Gerard Pons-Moll, and Yebin Liu.
\newblock Doublefusion: Real-time capture of human performances with inner body shapes from a single depth sensor.
\newblock \emph{TPAMI}, 2019.

\bibitem[Zanfir et~al.(2020)Zanfir, Bazavan, Zanfir, Freeman, Sukthankar, and Sminchisescu]{zanfir2020neural}
Andrei Zanfir, Eduard~Gabriel Bazavan, Mihai Zanfir, William~T Freeman, Rahul Sukthankar, and Cristian Sminchisescu.
\newblock Neural descent for visual 3d human pose and shape.
\newblock \emph{arXiv preprint arXiv:2008.06910}, 2020.

\bibitem[Zhang(2000)]{zhang2000flexible}
Zhengyou Zhang.
\newblock A flexible new technique for camera calibration.
\newblock \emph{IEEE Transactions on pattern analysis and machine intelligence}, 22\penalty0 (11):\penalty0 1330--1334, 2000.

\bibitem[Zhu et~al.(2020)Zhu, Ma, Wang, Xu, Shi, and Lin]{zhu2020ssn}
Xinge Zhu, Yuexin Ma, Tai Wang, Yan Xu, Jianping Shi, and Dahua Lin.
\newblock Ssn: Shape signature networks for multi-class object detection from point clouds.
\newblock In \emph{ECCV}, pages 581--597. Springer, 2020.

\bibitem[Zhu et~al.(2021)Zhu, Zhou, Wang, Hong, Li, Ma, Li, Yang, and Lin]{zhu2021cylindrical}
Xinge Zhu, Hui Zhou, Tai Wang, Fangzhou Hong, Wei Li, Yuexin Ma, Hongsheng Li, Ruigang Yang, and Dahua Lin.
\newblock Cylindrical and asymmetrical 3d convolution networks for lidar-based perception.
\newblock \emph{TPAMI}, 2021.

\end{thebibliography}
